\documentclass[11pt,a4paper]{article}
\usepackage[T1]{fontenc}
\usepackage[hyperref]{acl2020}
\usepackage{times}
\usepackage{latexsym}
\usepackage{enumitem}
\usepackage{amssymb}
\usepackage{amsthm}
\usepackage{amsmath}
\usepackage{color,soul}

\usepackage{graphicx}
\usepackage{subcaption}
\usepackage[disable]{todonotes}
\usepackage{multirow}
\usepackage{hyperref}
\usepackage{bm}
\usepackage{textcomp}

\usepackage{booktabs}
\usepackage{microtype}

\aclfinalcopy %
\def\aclpaperid{1114} %

\usepackage{comment}

\newcommand\yb[1]{\todo{\footnotesize YB: #1}}

\newcommand{\contextualizer}{f}
\newcommand{\wordrepr}{\mathbf{h}}

\newcommand{\attnhead}{\bm{\alpha}}

\newcommand{\ones}{\mathbf{1}} %
\newcommand{\bmat}[1]{\mathbf{#1}} %
\newcommand{\simmeas}{\texttt{sim}}
\newcommand{\svsim}{\texttt{svsim}}
\newcommand{\pwsim}{\texttt{pwsim}}
\newcommand{\neuronsim}{\texttt{neuronsim}}
\newcommand{\linckasim}{\texttt{ckasim}}
\newcommand{\mixedsim}{\texttt{mixedsim}}
\newcommand{\attnsim}{\texttt{attnsim}}
\newcommand{\norm}[1]{\ensuremath{\left\|#1\right\|}}
\newcommand{\rval}[1]{\ensuremath{#1\texttt{.r}}}

\DeclareMathOperator{\SVD}{SVD}
\DeclareMathOperator{\abs}{abs}
\DeclareMathOperator{\lstsq}{lstsq}
\DeclareMathOperator{\weights}{weights}
\DeclareMathOperator{\len}{len}
\DeclareMathOperator{\Sim}{Sim}
\DeclareMathOperator{\KL}{KL}

\title{Similarity Analysis of Contextual Word Representation Models}

\author{John M. Wu\thanks{~Equal contribution}~\textsuperscript{1} \hspace{2em} Yonatan Belinkov\textsuperscript{*12} \\ \textbf{Hassan Sajjad}\textsuperscript{3} \hspace{2em} \textbf{Nadir Durrani}\textsuperscript{3} \hspace{2em} \textbf{Fahim Dalvi}\textsuperscript{3} \hspace{2em} \textbf{James Glass}\textsuperscript{1} \\
\textsuperscript{1}MIT Computer Science and Artificial Intelligence Laboratory, Cambridge, MA, USA \\ 
\textsuperscript{2}Harvard John A. Paulson School of Engineering and Applied Sciences, Cambridge, MA, USA \\ 
\textsuperscript{3}Qatar Computing Research Institute, HBKU Research Complex, Doha 5825, Qatar \\ 
  \texttt{\{johnmwu,belinkov,glass\}@csail.mit.edu} \\ 
  \texttt{\{hsajjad,ndurrani,faimaduddin\}@qf.org.qa} \\}

\date{}

\begin{document}
\maketitle

\setlength{\abovedisplayskip}{3pt}
\setlength{\belowdisplayskip}{3pt}

\begin{abstract}

This paper investigates contextual word representation models from the lens of similarity analysis. Given a collection of trained models, we measure the similarity of their internal representations and attention. 
Critically, these models  
come from  vastly different architectures. We use existing and novel similarity measures that aim to gauge the level of localization of information in the deep models, and facilitate the investigation of which design factors affect model similarity, without requiring any external linguistic annotation.  The analysis reveals that  models within the same family are more similar to one another, as may be expected. Surprisingly, different architectures 
have rather similar representations, but different individual neurons. 
We also observed differences in information localization in lower and higher layers and found that higher layers are more affected by fine-tuning on downstream tasks.%
\footnote{The code is available at \url{https://github.com/johnmwu/contextual-corr-analysis}.
}

\end{abstract}

\section{Introduction}

Contextual word representations such as 
ELMo~\citep{peters2018deep} and BERT~\citep{devlin-etal-2019-BERT} have led to impressive improvements in a variety of tasks. 
With this progress in breaking the state of the art, interest in the community has expanded to analyzing such models in an effort to illuminate their inner workings. 
A number of studies have analyzed the internal representations in such models and attempted to assess what linguistic properties they capture. A prominent methodology for this is to train supervised classifiers based on the models' learned representations, and predict various linguistic properties.  
For instance, \citet{liu-etal-2019-linguistic} train such classifiers on 16 linguistic tasks, including part-of-speech tagging, chunking, named entity recognition, %
and others. Such an approach may reveal how well representations from different models, and model layers, capture different properties. This approach, known as analysis by probing classifiers, has been used in numerous other studies~\cite{belinkov-glass-2019-analysis}. 

While the above approach yields compelling insights, its applicability is constrained by the availability of linguistic annotations. In addition, comparisons of different models are indirect, via the probing accuracy, making it difficult to comment on the similarities and differences of different models. 
In this paper, we develop complementary methods for analyzing contextual word representations based on their inter- and intra-similarity. 
While this similarity analysis does not tell us absolute facts about a model, it allows comparing representations without subscribing to one type of information.
We consider several kinds of similarity measures based on different levels of localization/distributivity of information: from neuron-level pairwise comparisons of individual neurons to representation-level comparisons of full word representations. 
We also explore similarity measures based on models' attention weights, in the case of Transformer models~\cite{vaswani2017}. 
This approach enables us to ask questions such as: Do different models behave similarly on the same inputs? Which design choices determine whether models behave similarly or differently? Are certain model components more similar than others across architectures? Is the information in a given model more or less localized (encoded in individual components) compared to other models?\footnote{%
\citet{hinton1984distributed} defines a localist representation as one using one computing element for each represented entity. In a language model, this definition would depend on what linguistic concepts we deem important, and is thus somewhat arbitrary. We develop a measure that aims to capture this notion of localization without recourse to a specific set of linguistic properties.} 

We choose a collection of pre-trained models that aim to capture diverse aspects of modeling choices, including the building blocks (Recurrent Networks, Transformers), language modeling objective  (unidirectional, bidirectional, masked, permutation-based), and model depth (from 3 to 24 layers). 
More specifically, we experiment with variants of ELMo, BERT, GPT~\cite{radford2018improving}, GPT2~\cite{radford2019language}, and XLNet~\cite{yang2019xlnet}. 
Notably, we use the same methods to investigate the effect that  fine-tuning on downstream tasks has 
on the model similarities. %

Our analysis yields the following insights:
\begin{itemize} %
    \item Different architectures may have similar representations, but different individual neurons. Models within the same family are more similar to one another in terms of both their neurons and full representations.
    \item Lower layers are more similar than higher layers across architectures. 
    \item  Higher layers have more localized representations than lower layers. 
    \item Higher layers are more affected by fine-tuning than lower layers, in terms of their representations and attentions, and thus are less similar to the higher layers of pre-trained models. 
    \item Fine-tuning affects the localization of information, causing high layers to be less localized. 
\end{itemize}

Finally, we show how the similarity analysis can motivate a simple technique for efficient fine-tuning, where freezing the bottom layers of models still maintains comparable performance to fine-tuning the full network, while reducing the fine-tuning time.

\section{Related Work}

The most common approach for analyzing neural network models in general, and contextual word representations in particular, is by probing classifiers 
\cite{ettinger-etal-2016-probing,belinkov:2017:acl,adi2017fine,conneau-etal-2018-cram,hupkes2018visualisation}, where a classifier is trained on a corpus of linguistic annotations using representations from the model under investigation. For example, \citet{liu-etal-2019-linguistic} used this methodology for investigating the representations of contextual word representations on 16 linguistic tasks. 
One limitation of this approach is that it requires specifying linguistic tasks of interest and obtaining suitable annotations. This potentially limits the applicability of the approach. 

An orthogonal analysis method relies on similarities between model representations. \citet{bau:2019:ICLR} used this approach to analyze the role of individual neurons in neural machine translation. %
They found that individual neurons are important and interpretable. However, their work was limited to a certain kind of architecture (specifically,  a recurrent one). In contrast, we %
compare models of various architectures and objective functions. 

Other work used similarity measures to study learning dynamics in language models by comparing checkpoints of recurrent language models~\cite{NIPS2018_7815}, or a language model and a part-of-speech tagger~\cite{saphra-lopez-2019-understanding}. 
Our work adopts a similar approach, but explores a range of similarity measures over  different contextual word representation models. 

Questions of localization and distributivity of information %
have been under investigation for a long time in the connectionist cognitive science literature~\cite{page_2000,BOWERS2002413,doi:10.1080/09540091.2011.587505}. %
While neural language representations are thought to be densely distributed, several recent studies have %
pointed out the importance of individual neurons~\cite{qian-etal-2016-analyzing,shi-etal-2016-neural,radford2017learning,lakretz-etal-2019-emergence,bau:2019:ICLR,dalvi2019one,baan-etal-2019-realization}. Our study contributes to this line of work by designing measures of localization and distributivity of information in a collection of models. Such measures may facilitate incorporating neuron interactions in new training objectives~\cite{li2020neuron}.  

\section{Similarity Measures}

We present five groups of similarity measures, each capturing a different similarity notion. Consider a collection of $M$ models $\{\contextualizer^{(m)}\}_{m=1}^M$, yielding word representations $\wordrepr^{(m)}_l$ and potentially attention weights $\attnhead_{l}^{(m)}$ at each layer $l$. Let $k$ index neurons $\wordrepr^{(m)}_l[k]$ or attention heads $\attnhead_{l}^{(m)}[k]$. $\wordrepr^{(m)}_l[k]$, $\attnhead_{l}^{(m)}[k]$ are real (resp.\ matrix) valued, ranging over words (resp.\ sentences) in a corpus. %
Our similarity measures are of the form $\simmeas(\wordrepr^{(m)}_l, \wordrepr^{(m')}_{l'})$ or $\simmeas(\attnhead^{(m)}_{l}, \attnhead^{(m')}_{l'})$, that is, they find similarities between layers. %
We present the full mathematical details in appendix~\ref{app:math}. %

\subsection{Neuron-level similarity}

A neuron-level similarity measure captures %
similarity between pairs of individual neurons. We consider one such measure, $\neuronsim$, following \citet{bau:2019:ICLR}. 
For every neuron $k$ in layer $l$,  $\neuronsim$~finds the maximum correlation between it and another neuron in another layer $l'$. Then, it averages over neurons in layer $l$.\footnote{In this and other measures that allowed it, we also experimented with averaging just the top $k$ neurons (or canonical correlations, in Section~\ref{sec:repsim} measures) in case most of the layer is noise. Heatmaps are in the online repository. We did not notice major differences.} 
This measure aims to capture localization of information. It is high when two layers have pairs of neurons with similar behavior. This is far more likely when the models have local, rather than distributed representations, because for distributed representations to have similar pairs of neurons the information must be distributed similarly. %

\subsection{Mixed neuron--representation similarity}
A mixed neuron--representation similarity measure captures a similarity between a neuron in one model with a layer in another. We consider one such measure, $\mixedsim$:
for every neuron $k$ in layer $l$, regress to it from all neurons in layer $l'$ and measure the quality of fit. Then, average over neurons in $l$.
It is possible that some information is localized in one layer but distributed in another layer. $\mixedsim$ captures such a phenomenon.

\subsection{Representation-level similarity} \label{sec:repsim}
 A representation-level measure finds correlations between a full model (or layer) simultaneously. We consider three such measures: two based on canonical correlation analysis (CCA), namely singular vector CCA (\svsim; \citealt{NIPS2017_7188}) and projection weighted CCA (\pwsim; \citealt{NIPS2018_7815}), in addition to linear centered kernel alignment (\linckasim; \citealt{pmlr-v97-kornblith19a}).\footnote{We also experimented with the RBF variant, which is computationally demanding. We found similar patterns in preliminary experiments, so we focus on the linear variant.}
  These measures emphasize distributivity of information---if two layers behave similarly over all of their neurons, the similarity will be higher, even if no individual neuron has a similar matching pair or is represented well by all neurons in the other layer. 
 
 Other representation-level similarity measures may be useful, such as representation similarity analysis (RSA; \citealt{10.3389/neuro.06.004.2008}), which has been used to analyze neural network representations \cite{bouchacourt-baroni-2018-agents,chrupala-alishahi-2019-correlating,chrupala-2019-symbolic}, or other variants of CCA, such as deep CCA~\cite{pmlr-v28-andrew13}. We leave the explorations of such measures to future work.

\subsection{Attention-level similarity}
Previous work analyzing network similarity has mostly focused on representation-based similarities~\cite{NIPS2018_7815,saphra-lopez-2019-understanding,voita-etal-2019-bottom}. Here we consider similarity based on attention weights in Transformer models. 

Analogous to a neuron-level similarity measure, an attention-level similarity measure finds the most ``correlated'' other attention head. We consider three methods to correlate heads, based on the norm of two attention matrices  $\attnhead_{l}^{(m)}[k]$, $\attnhead^{(m')}_{l'}[k']$,
their Pearson correlation, and their Jensen--Shannon divergence.\footnote{Other recent work has used the Jensen--Shannon divergence to measure distances between attention heads \cite{clark-etal-2019-bert,jain-wallace-2019-attention}.} %
We then average over heads $k$ in layer $l$, as before. 
These measures are similar to %
\texttt{neuronsim} %
in that they emphasize localization of information---if two layers have pairs of heads that are very similar in their behavior, the similarity will be higher.

\subsection{Distributed attention-level similarity}
We consider parallels of the representation-level similarity. To compare the entire attention heads in two layers, we concatenate all weights from all heads in one layer to get an attention representation.  That is, we obtain attention representations $\attnhead^{(m)}_l[h]$, a random variable ranging over pairs of words in the same sentence, such that $\attnhead^{(m)}_{l,(i,j)}[h]$ is a scalar value. It is a matrix where the first axis is indexed by word pairs, and the second by heads. 
We flatten these matrices and  use \svsim, \pwsim, and \linckasim~as above for comparing these attention representations. These measures should be high when the entire set of heads in one layer is similar to the set of heads in another layer.

 \vspace{-5pt}
\section{Experimental Setup}

\begin{figure*}[t]
    \centering
    \begin{subfigure}[b]{0.49\linewidth}
    \centering
    \includegraphics[width=\linewidth]{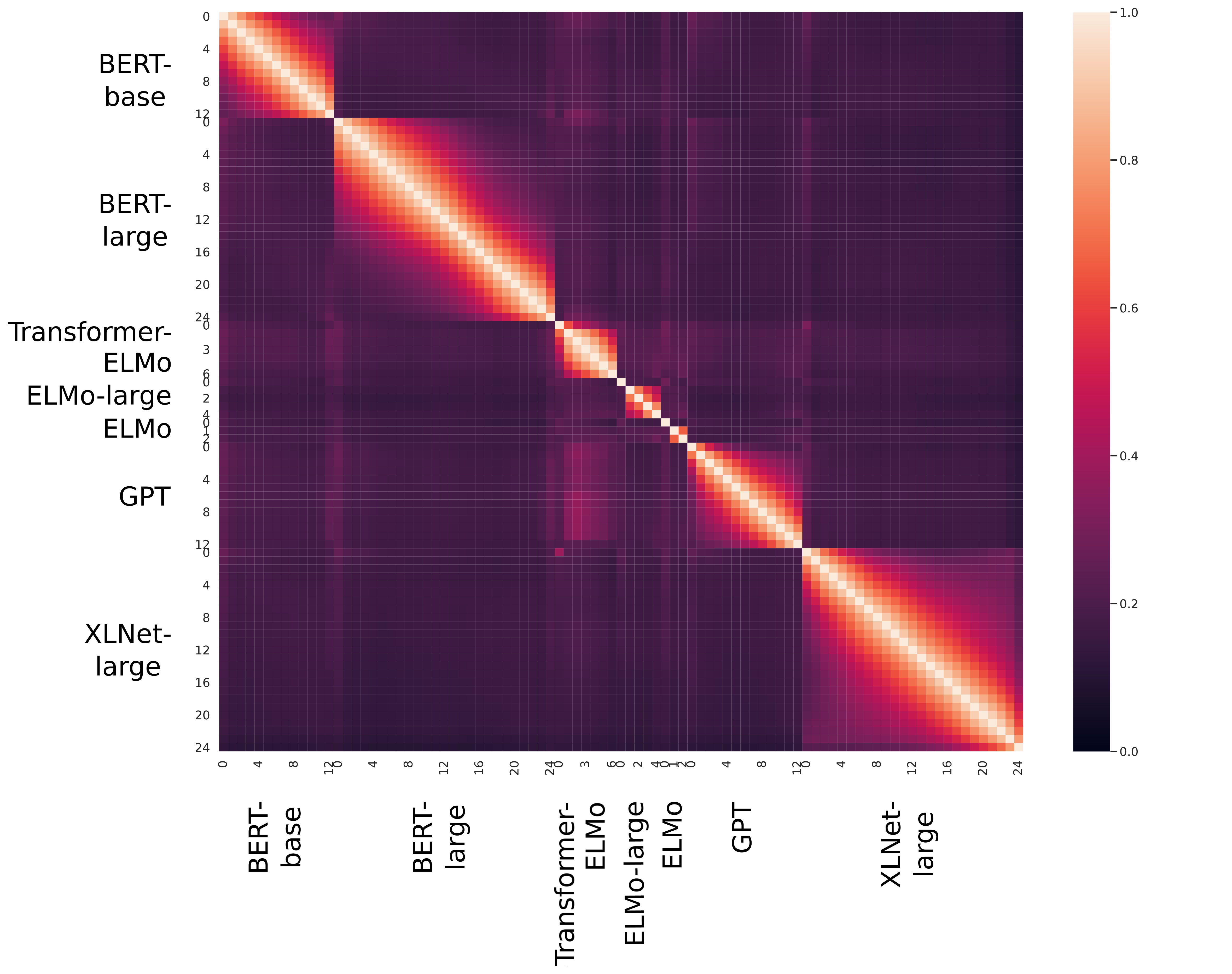}
    \caption{\neuronsim}
    \label{fig:heatmap-maxcorr}
    \end{subfigure}
    \begin{subfigure}[b]{0.49\linewidth}
    \centering
    \includegraphics[width=\linewidth]{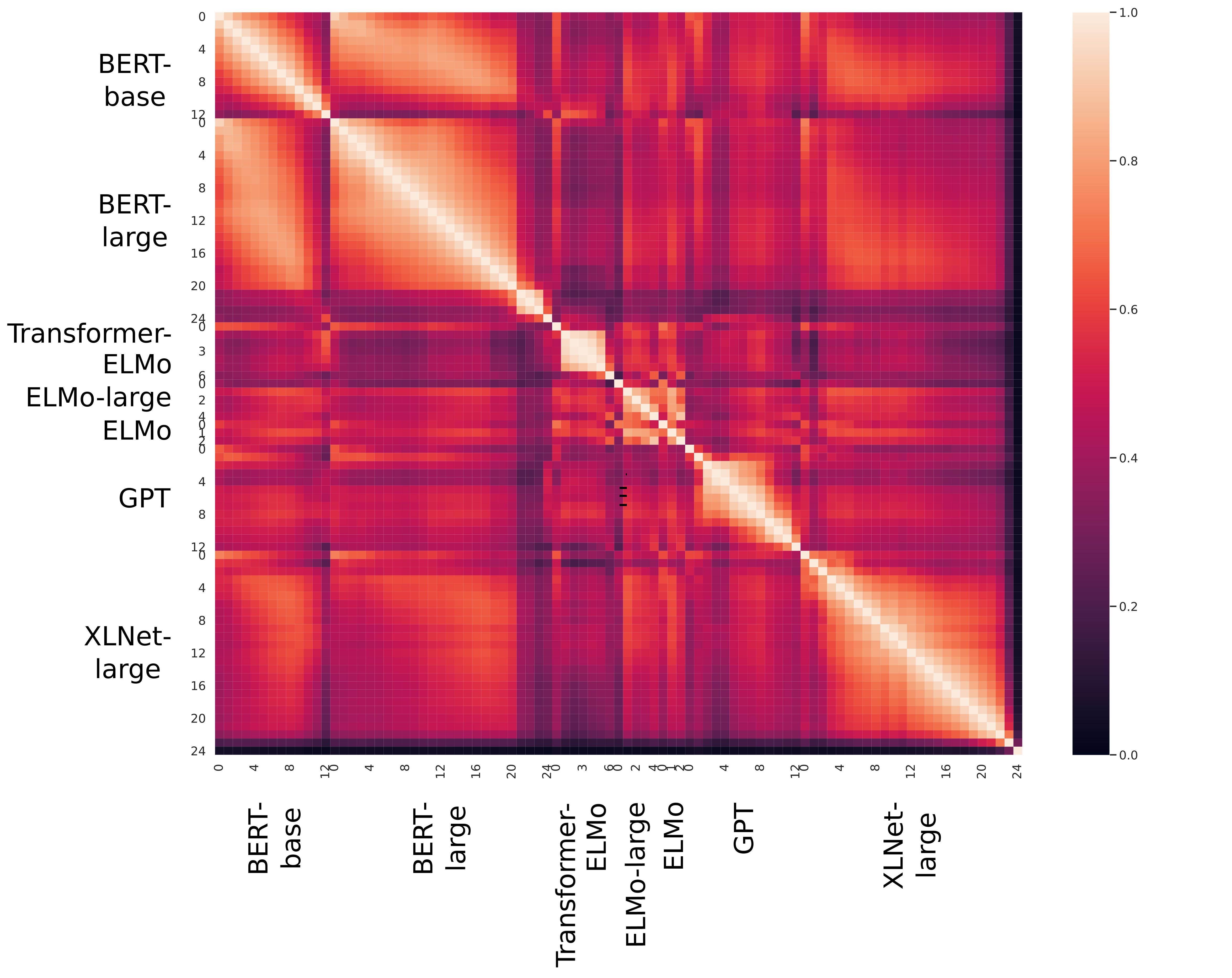}
    \caption{\linckasim}
    \label{fig:heatmap-cka-main}
    \end{subfigure}    
    \caption{Similarity heatmaps  of layers in various models under %
    neuron- %
    and representation-level similarities. %
    }
    \label{fig:heatmaps}
\end{figure*}

\paragraph{Models}
We choose a collection of pre-trained models that aim to capture diverse aspects of modeling choices, including the building blocks (RNNs, Transformers), language modeling objective  (unidirectional, bidirectional, masked, permutation-based), and model depth (from 3 to 24 layers).

\vspace{-3pt}
\paragraph{\it ELMo variants}
We use the original ELMo \cite{peters2018deep}, a bidirectional RNN model with two hidden layers, as well as two variants -- a deeper and larger 4-layer model and a Transformer-equivalent variant %
\yb{change to ``Transformer-ELMo'', also in figures}  \cite{peters-etal-2018-dissecting}.  

\vspace{-3pt}
\paragraph{\it  GPT variants} We use both the original OpenAI Transformer (GPT;~\citealt{radford2018improving}) and its successor GPT2~\cite{radford2019language}, in the small and medium model sizes.  These are all unidirectional Transformer LMs. 

\vspace{-3pt}
\paragraph{\it BERT} We use BERT-base/large (12/24 layers;~\citealt{devlin-etal-2019-BERT}):  Transformer LMs trained with a masked LM objective function.\footnote{BERT is also trained with a next sentence prediction objective, although this may be redundant \cite{liu2019roBERTa}.}

\vspace{-3pt}
\paragraph{\it XLNet} We use XLNet-base/large (12/24 layers;~\citealt{yang2019xlnet}). Both are Transformer LM with a permutation-based objective function. %

\paragraph{Data}
For analyzing the models, we run them on the Penn Treebank development set \cite{marcus-etal-1993-building}, following the setup taken by \citet{liu-etal-2019-linguistic} in their probing classifier experiments.\footnote{As suggested by a reviewer, we verified that the results are consistent when using another dataset (Appendix~\ref{app:altdata}).} We collect representations and attention weights from each layer in each model for computing the similarity measures.  We obtain representations for models used in \citet{liu-etal-2019-linguistic} from their implementation and use the transformers library \cite{Wolf2019HuggingFacesTS} to extract other representations. 
We aggregate sub-word representations by taking the representation of the last sub-word, following \citet{liu-etal-2019-linguistic}, and sub-word attentions by summing up attention to sub-words and averaging attention from sub-words, following \citet{clark-etal-2019-bert}, which guarantees that the attention from each word sums to one.

\section{Similarity of Pre-trained Models}

\subsection{Neuron and representation levels } 
\label{sec:results-repr}

Figure~\ref{fig:heatmaps} shows heatmaps of similarities between layers of different models, according to 
\neuronsim~and \linckasim. Heatmaps for the other measures are provided in Appendix~\ref{app:representation}. 
The heatmaps reveal the following insights. 

\paragraph{Different architectures may have similar representations, but different individual neurons}

Comparing the heatmaps, the most striking distinction is %
that 
\neuronsim~induces a distinctly block-diagonal heatmap, reflecting high intra-model similarities and low inter-model similarities. As \neuronsim~is computed by finding pairs of very similar neurons, this means that within a model, different layers have similar individual neurons, but across models, neurons are very different. 
In contrast, \linckasim - show fairly significant similarities across models (high values off the main diagonal), 
indicating that different models generate similar representations.
The most similar cross-model similarities are found by \mixedsim~(Figure~\ref{fig:heatmap-linreg} in Appendix~\ref{app:representation}), which suggests that individual neurons in one model may be well represented by a linear combination of neurons in another layer. 
The other representation-level similarities (\linckasim, \svsim, and \pwsim), also show cross-model similarities, albeit to a lesser extent.

\paragraph{Models within the same family are more similar}
The heatmaps show greater similarity within a model than across models (bright diagonal). 
Different models sharing the same architecture and objective function, but different depths, also exhibit substantial representation-level similarities -- for instance, compare BERT-base and BERT-large or ELMo-original and ELMo-4-layers, under \linckasim~(Figure~\ref{fig:heatmap-cka-main}). 
The Transformer-ELMo %
presents an instructive case, as it shares ELMo's bidirectional objective function but with Transformers rather than RNNs. Its layers are mostly similar to themselves and the other ELMo models, but also to GPT, more so than to BERT or XLNet, which use masked and permutation language modeling objectives, respectively. 
Thus it seems that the objective has a considerable impact on representation similarity.\footnote{\citet{voita-etal-2019-bottom} found that differences in  the training objective result in more different representations  (according to \pwsim) than differences in random initialization.}  %

\begin{figure}[t]
    \centering
    \includegraphics[width=\linewidth]{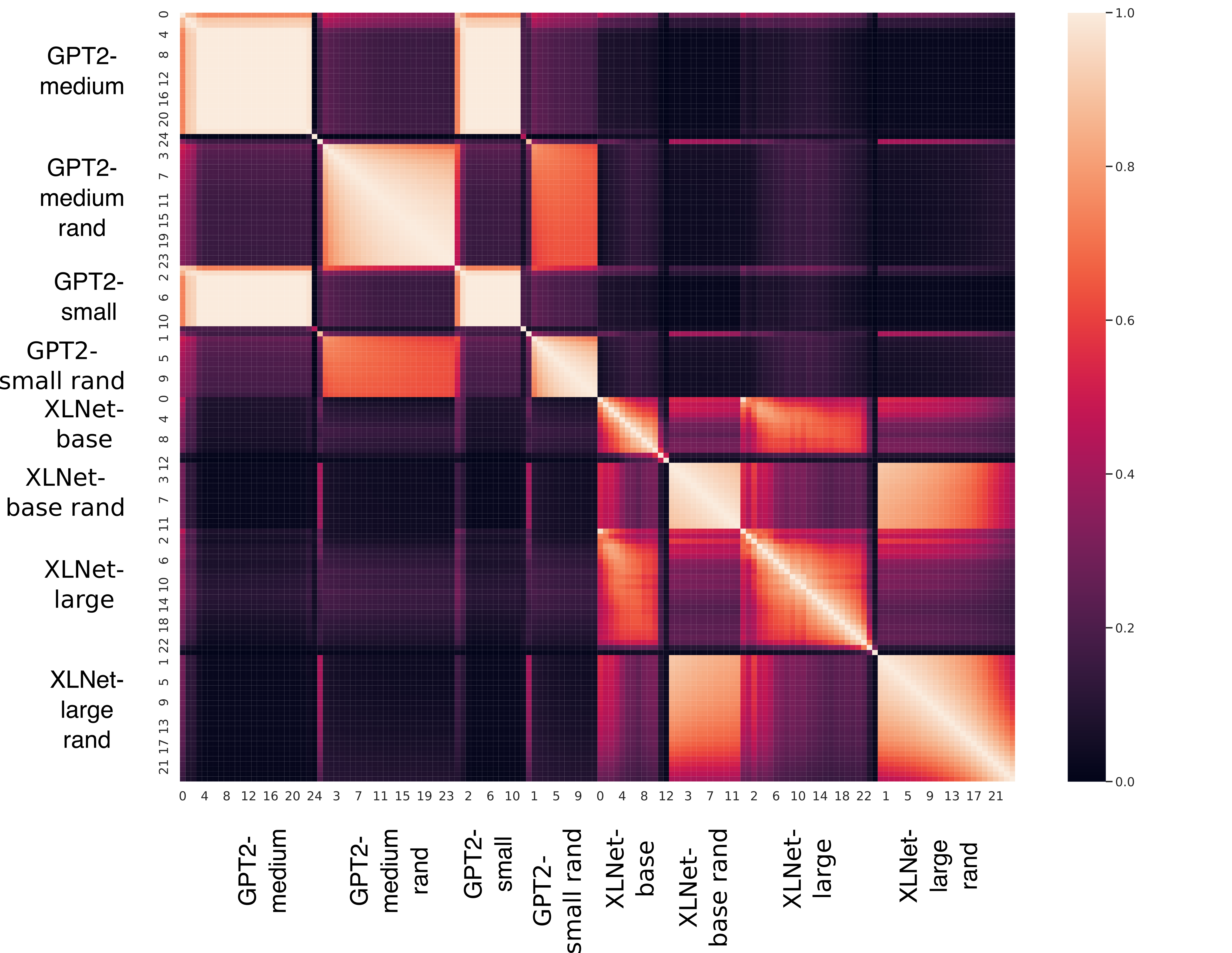}
    \caption{ %
    \linckasim~similarity heatmap  of layers in base and random models.}
    \label{fig:random}
\end{figure}

The fact that models within the same family are more similar to each other supports the choice of \citet{saphra-lopez-2019-understanding} to use models of similar architecture when probing models via similarity measures across tasks.\footnote{We thank a reviewer for pointing out this connection.}
A possible confounder is that models within the same family are trained on the same data, but cross-family models are trained on different data. It is difficult to control for this given the computational demands of training such models  and the current practice in the community of training models on ever increasing sizes of data, rather than a standard fixed dataset. 
However, Figure~\ref{fig:random} shows similarity heatmaps %
of layers from pre-trained and randomly initialized models using  %
\linckasim, exhibiting  %
high intra-model similarities, as before. Interestingly, models within the same family  (either GPT2 or XLNet) are more similar than across families, even with random models, indicating that intrinsic aspects of models in a given family make them similar, regardless of the training data or process.\footnote{Relatedly, \citet{NIPS2018_7815} found similar CCA coefficients in representations from recurrent language models trained on different datasets.} 
As may be expected, in most cases, the similarity between random and pre-trained models is small. One exception is the vertical bands in the lower triangle, which indicate that the bottom layers of trained models are similar to many layers of random models. This may be due to random models merely transferring information from bottom to top, without meaningful processing. 
Still, it may explain why random models sometimes generate useful features \cite{wieting2018no}. 
Meanwhile, as pointed out by a reviewer, lower layers converge faster, leaving them closer to their initial random state~\cite{NIPS2017_7188,shwartz2017opening}.

\paragraph{Lower layers are more similar across architectures} 
The representation-level heatmaps (Figure \ref{fig:heatmaps}) all exhibit horizontal stripes at lower layers, especially with \linckasim, indicating that lower layers are more similar than higher layers when comparing across models. This pattern can be explained by lower layers being closer to the input, which is always the same words.   A similar observation has been made for vision networks  \cite{NIPS2017_7188}.\footnote{\citet{NIPS2017_7188} also used \svsim~to study recurrent language models, showing that lower layers converge faster. Although they have not looked at cross-model comparisons, faster convergence may be consistent with fewer changes during training, which can explain why lower layers are more similar across architectures.} 
\citet{voita-etal-2019-bottom} found a similar pattern comparing Transformer models with different objective functions.

\vspace{-4pt}
\paragraph{Adjacent layers are more similar}
All heatmaps in Figure \ref{fig:heatmaps} exhibit a very bright diagonal and bright lines slightly off the main diagonal, indicating that adjacent layers are more similar. This is even true when 
comparing layers of different models (notice the diagonal nature of BERT-base vs.\ BERT-large in Figure~\ref{fig:heatmap-cka-main}),  %
indicating that layers at the same relative depth are more similar than layers at different relative depths.  
A similar pattern was found %
in vision networks~\cite{pmlr-v97-kornblith19a}.
Some patterns %
are unexpected. For instance, comparing XLNet with the 
BERT models, it appears that lower layers of XLNet are more similar to higher layers of BERT. %
We speculate that this is an artifact of the permutation-based objective in XLNet. 

We found corroborating evidence for this observation in ongoing parallel work, where we compare BERT and XLNet at different layers through word- \cite{liu-etal-2019-linguistic} and sentence-level tasks \cite{wang2019glue}:  while BERT requires mostly features from higher layers to achieve state-of-the-art results, in XLNet lower and middle layers suffice.

\begin{figure}[t]
    \centering
    \includegraphics[width=1\linewidth]{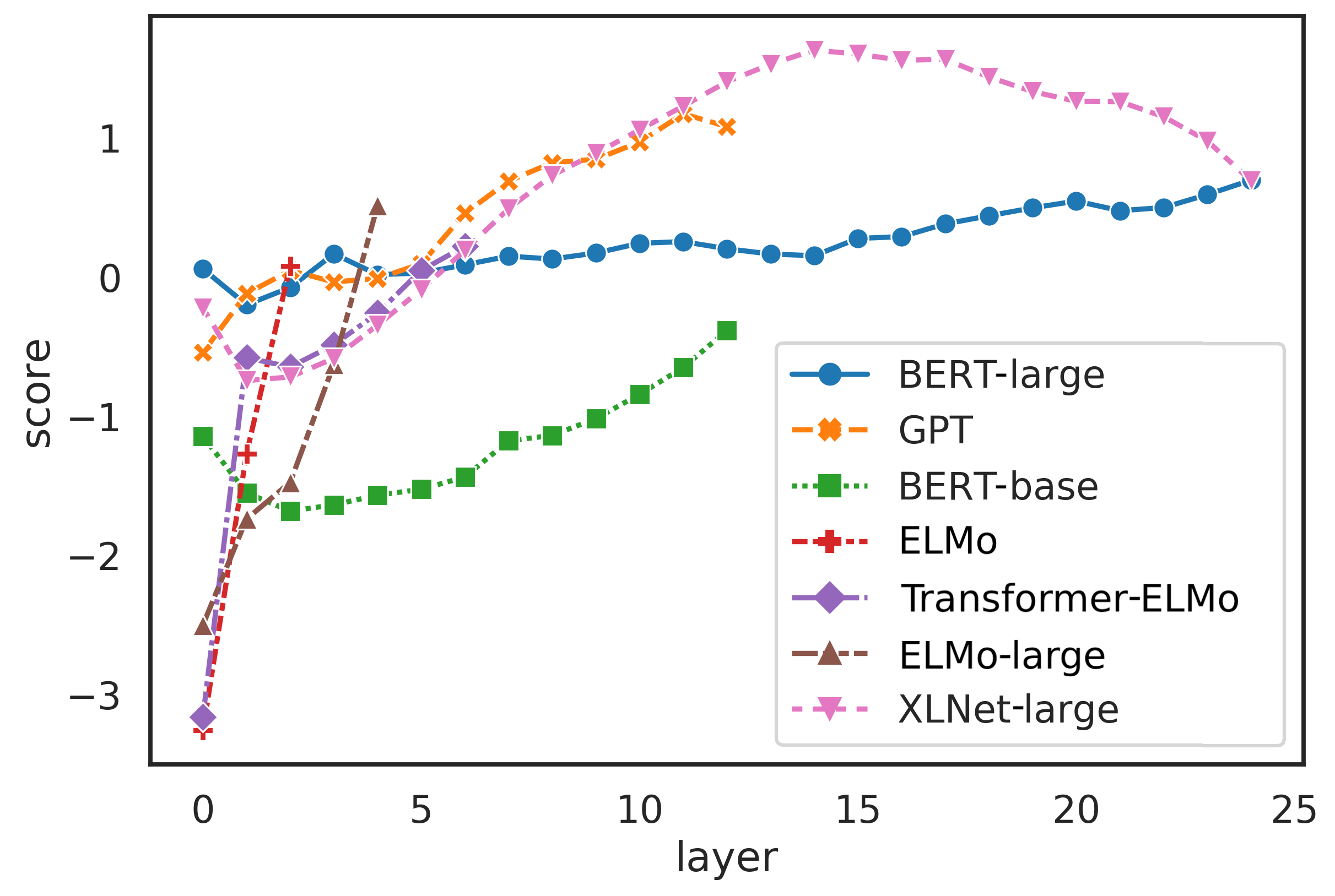}
    \caption{Localization score of various model layers.}
    \label{fig:all-loc-score}
\end{figure}

\paragraph{Higher layers are more localized than lower ones}
The different similarity measures capture different levels of localization vs.\ distributivity of information. \neuronsim~captures cases of localized information, where pairs of neurons in different layers behave similarly. \svsim~captures cases of distributed information, where the full layer representation is similar. 
To quantify these differences, we compute the average similarity according to each measure when comparing each layer to all other layers. In effect, we take the column-wise mean of each heatmap. We do this separately for \svsim~as the distributed measure and \neuronsim~as the localized measure, and we subtract the \svsim~means from the \neuronsim~means. This results in a measure of localization per layer. 
Figure~\ref{fig:all-loc-score} shows the results.

In all models, the localization score mostly increases with layers, indicating that information tends to become more localized at higher layers.\footnote{Recurrent models are more monotonous than Transformers, echoing results by~\citet{liu-etal-2019-linguistic} on language modeling perplexity in different layers.} 
This pattern is quite consistent, but may be surprising given prior observations on lower layers capturing phenomena that operate at a local context \cite{tenney-etal-2019-BERT}, which presumably require fewer neurons.
However, this pattern is in line with observations made by \citet{ethayarajh-2019-contextual}, who reported 
that upper layers of pre-trained models produce more context-specific representations. There appears to be a correspondence between our localization score and \citeauthor{ethayarajh-2019-contextual}'s context-specificity score, which is based on the cosine similarity of representations of the same word in different contexts. Thus, more localized representations are also more context-specific. A direct comparison between context-specificity and localization may be fruitful avenue for future work. 

Some models seem less localized than others, especially the 
ELMo variants, although this may be confounded by their being shallower models. BERT and XLNet models first decrease in localization and then increase. Interestingly, XLNet's localization score decreases towards the end, suggesting that its top layer representations are less context-specific.

\begin{figure*}[t]
    \centering
    \begin{subfigure}[b]{0.49\linewidth}
    \centering
    \includegraphics[width=\linewidth]{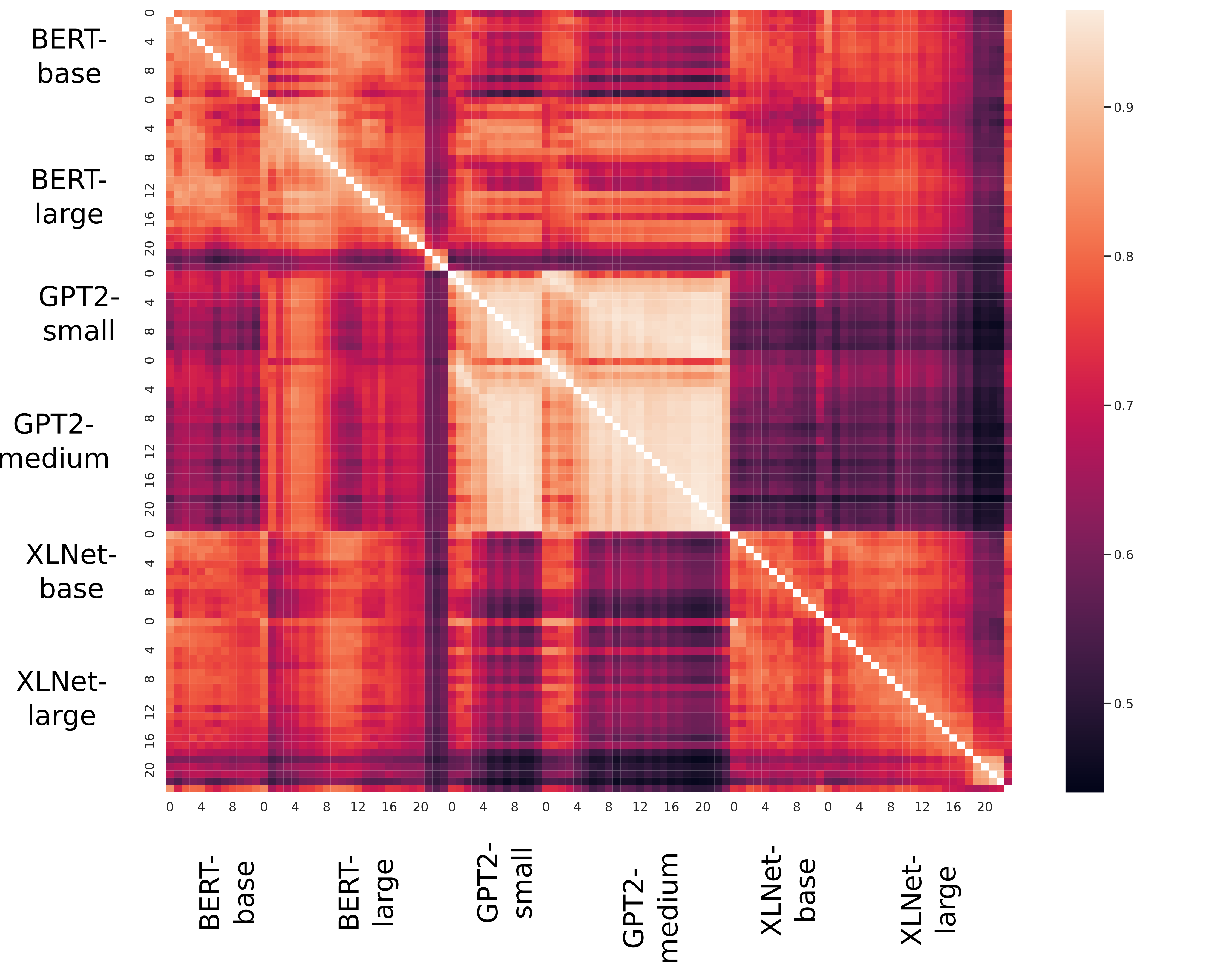}
    \caption{Jensen--Shannon}
    \label{fig:heatmap-attn-js-main}
    \end{subfigure}    
    \begin{subfigure}[b]{0.49\linewidth}
    \centering
    \includegraphics[width=\linewidth]{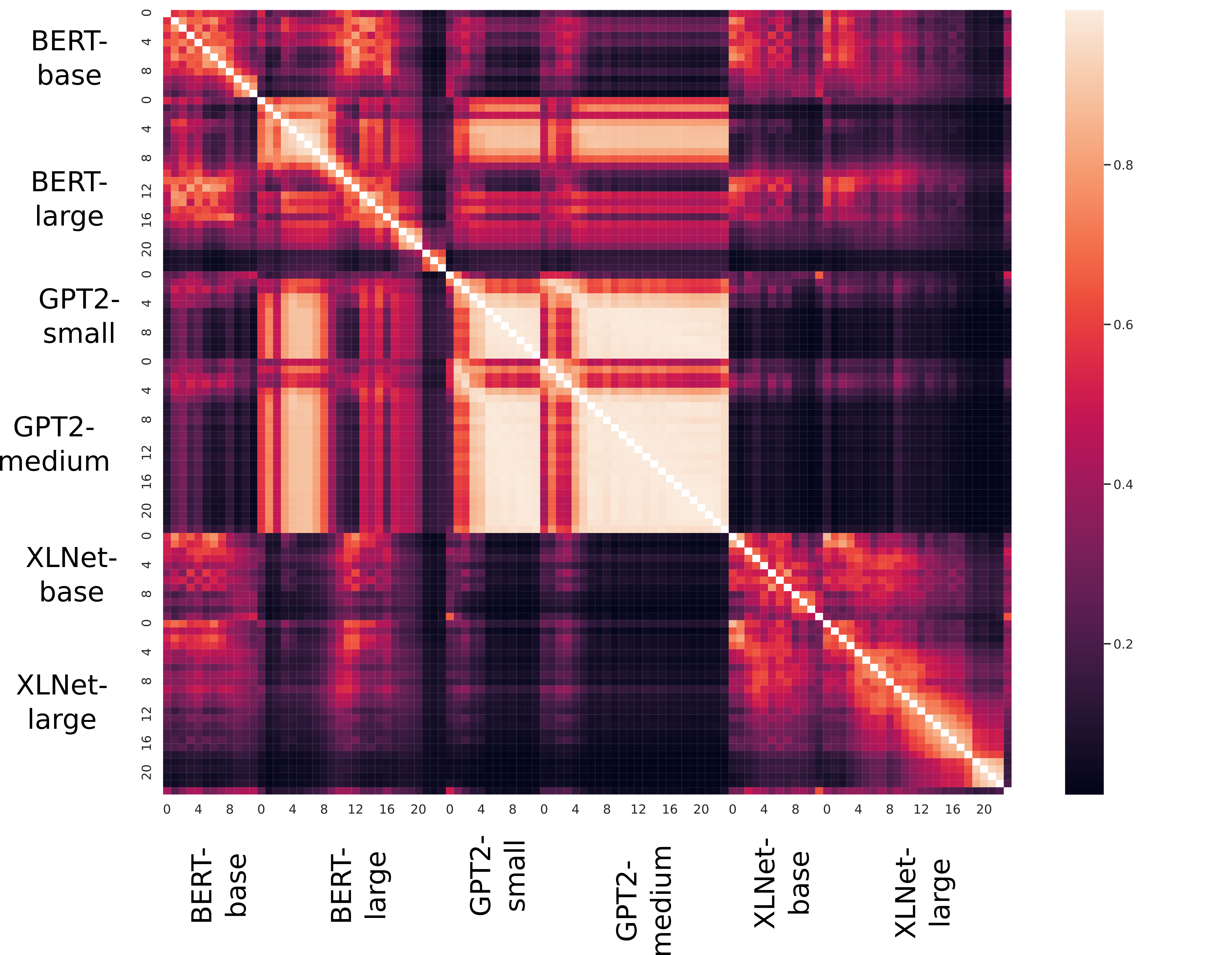}
    \caption{\linckasim}
    \label{fig:heatmap-attn-cka-main}
    \end{subfigure}  
    \caption{Similarity heatmaps  of layers in various models under two %
    attention-level similarity measures.}
    \label{fig:attn-heatmaps}
\end{figure*}

\subsection{Attention level} 

Figure~\ref{fig:attn-heatmaps} shows similarity heatmaps using two of the attention-level similarity measures---Jensen--Shannon and \linckasim---for layers from 6 models: BERT-base/large, GPT2-small/medium, and XLNet-base/large.
Layers within the same model or model family exhibit higher similarities (bright block diagonal), in line with results from the representation-level analysis. In particular, under both %
measures,  GPT2 layers are all very similar to each other, except for the bottom ones. 
Comparing the two heatmaps, the localized Jensen--Shannon similarity (Figure~\ref{fig:heatmap-attn-js-main}) shows higher similarities off the main diagonal than the distributed \linckasim~measure (Figure~\ref{fig:heatmap-attn-cka-main}), indicating that different models have pairs of attention heads that behave similarly, although the collection of heads from two different models is different in the aggregate. 
Heatmaps for the other %
measures are provided in Appendix~\ref{app:attention}, following primarily the same patterns. 

It is difficult to identify patterns within a given model family. However, under the attention-based \svsim~(Figure~\ref{fig:heatmap-attn-svcca} in Appendix~\ref{app:attention}), and to a lesser extent \pwsim~(Figure~\ref{fig:heatmap-attn-pwcca}), we see bright diagonals when comparing different GPT2 (and to a lesser extent XLNet and BERT) models, such that layers at the same relative depth are similar in their attention patterns. We have seen such a result also in the representation-based similarities. 

Adjacent layers seem more similar in some cases, but these patterns are often swamped by the large intra-model similarity. This result differs from our results for representational similarity. 

GPT2 models, at all layers, are similar to the bottom layers of BERT-large, expressed in bright vertical bands. In contrast, GPT2 models do not seem to be especially similar to XLNet. 
Comparing XLNet and BERT, we find that lower layers of XLNet are quite similar to higher layers of BERT-base and middle layers of BERT-large. This parallels the findings from comparing representations of XLNet and BERT,  which we conjecture is the result of the permutation-based objective in XLNet. %

In general, we find the attention-based similarities to be mostly in line with the neuron- and representation-level similarities. Nevertheless, they appear to be harder to interpret, as fine-grained patterns are less noticeable. 
One might mention in this context concerns regarding the reliability of attention weights for interpreting the importance of input words in a model~\cite{jain-wallace-2019-attention,serrano-smith-2019-attention,Brunner2020On}. However, characterizing the effect of such concerns on our attention-based similarity measures is beyond the current scope.

\section{Similarity of Fine-tuned Models}

How does fine-tuning on downstream tasks affect model similarity? 
In this section, we compare pre-trained models and their fine-tuned versions.
We use four of the GLUE tasks~\cite{wang2019glue}:  %

\vspace{-5pt}
\paragraph{MNLI} A multi-genre natural language inference dataset \cite{williams2018broad}, where the task is to predict whether a premise entails a hypothesis. 

\vspace{-5pt}
\paragraph{QNLI} A conversion of the Stanford question answering dataset~\cite{rajpurkar2016squad}, where the task is to determine whether a sentence contains the answer to a question. 

\vspace{-5pt}
\paragraph{QQP} A collection of question pairs from the Quora website, where the task is to determine whether two questions are semantically equivalent. 

\vspace{-5pt}
\paragraph{SST-2} A binary sentiment analysis task using the Stanford sentiment treebank~\cite{socher2013recursive}.

\begin{figure*}[t]
    \centering
    \begin{subfigure}[b]{0.49\linewidth}
    \centering
    \includegraphics[width=\linewidth]{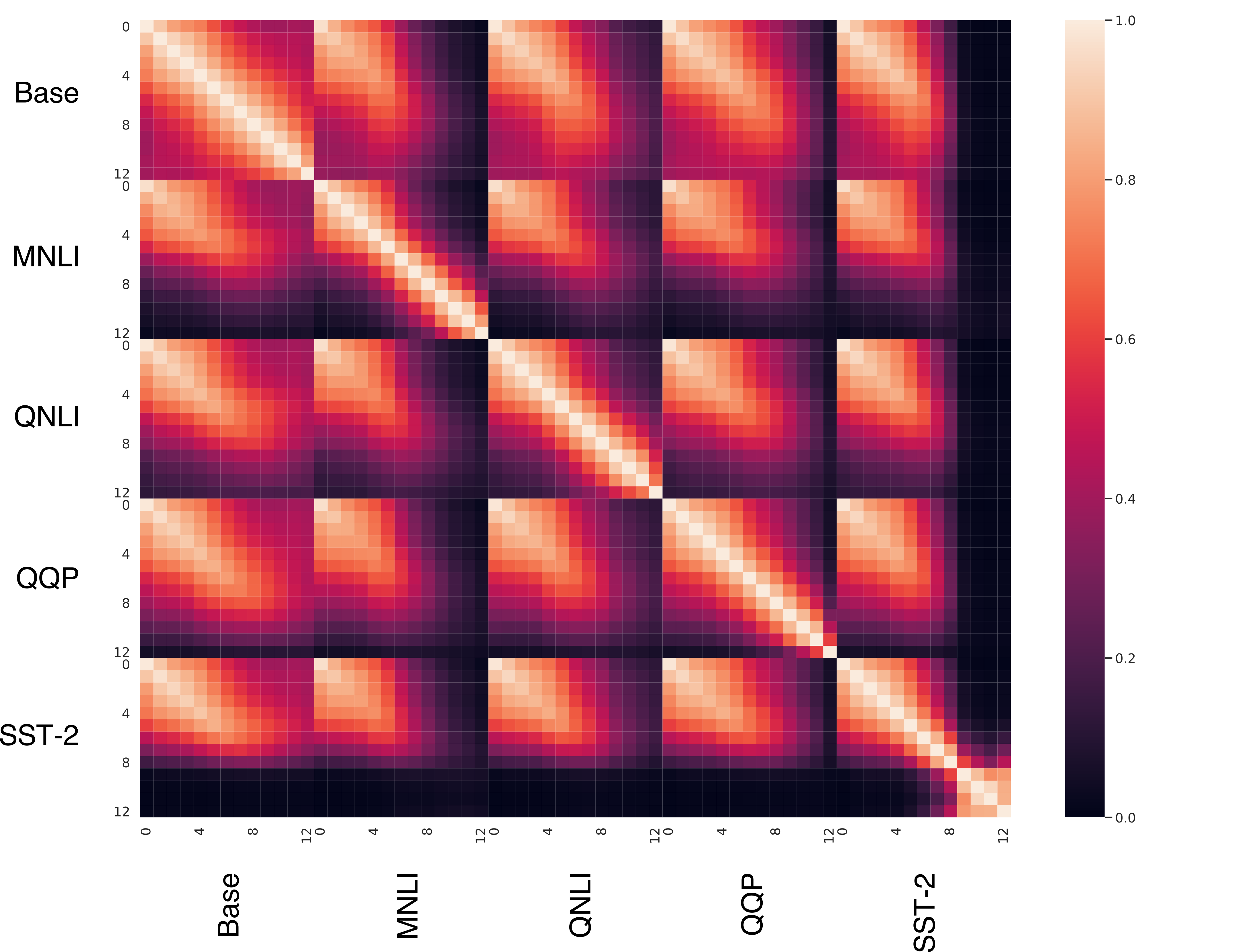}
    \caption{BERT}
    \label{fig:fine-tuned-heatmap-BERT}
    \end{subfigure}
    \begin{subfigure}[b]{0.49\linewidth}
    \centering
    \includegraphics[width=\linewidth]{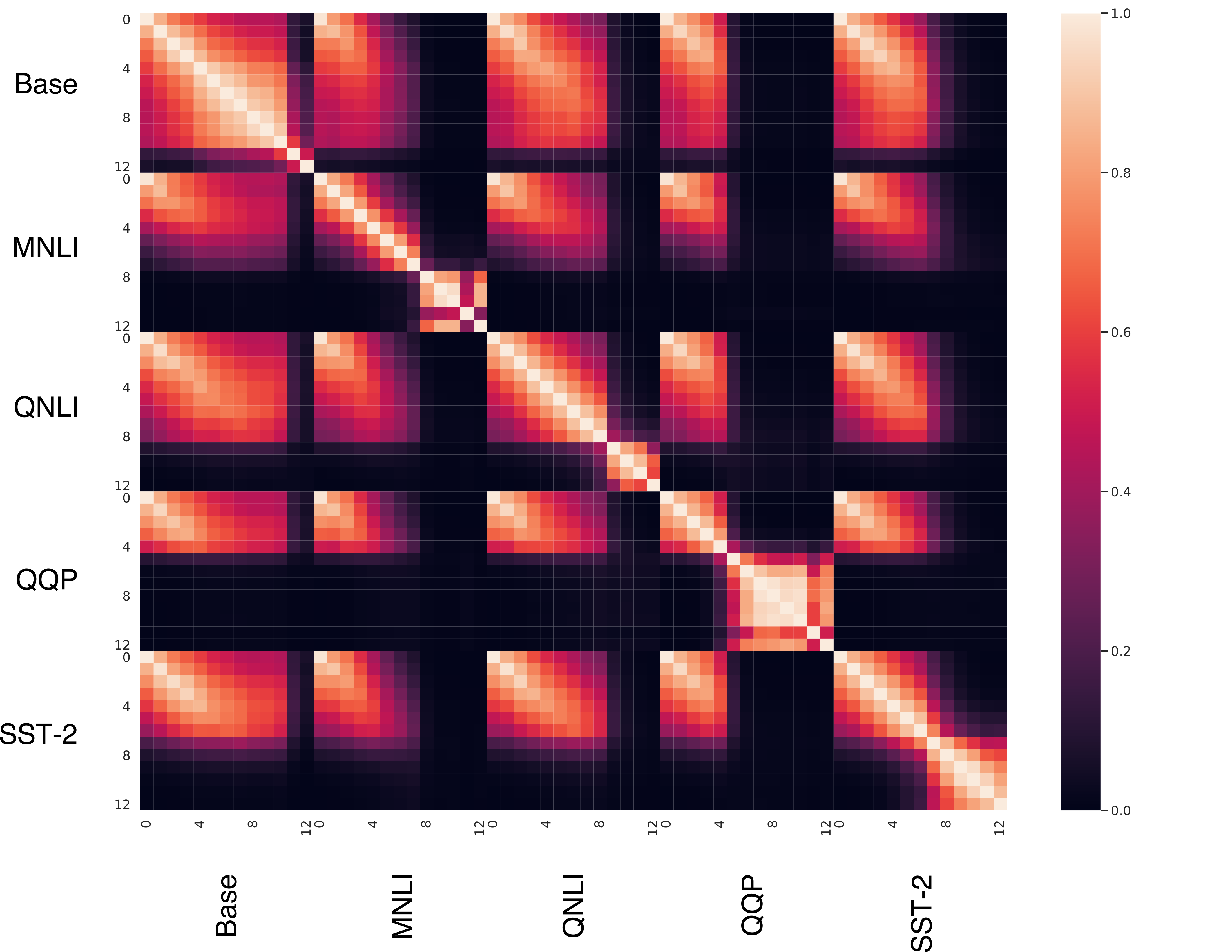}
    \caption{XLNet}
    \label{fig:fine-tuned-heatmap-xlnet}
    \end{subfigure}
    \caption{\linckasim~similarity heatmaps  of layers in base (pre-trained, not fine-tuned) and fine-tuned models.}
    \label{fig:fine-tuned-heatmaps}
\end{figure*}

\begin{figure*}[t]
    \centering
    \begin{subfigure}[b]{0.49\linewidth}
    \centering
    \includegraphics[width=\linewidth]{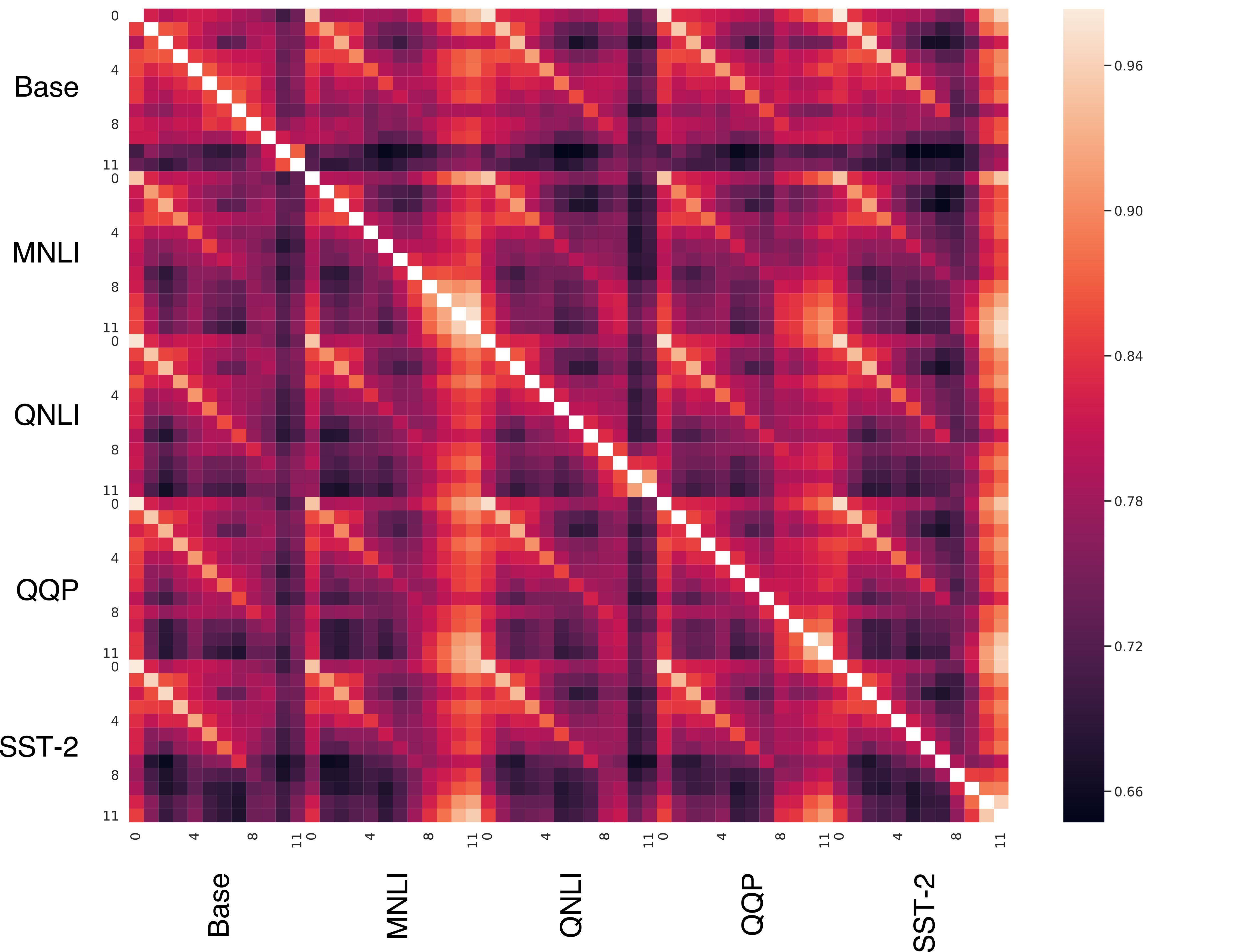}
    \caption{BERT}
    \label{fig:fine-tuned-attn-heatmap-BERT}
    \end{subfigure}
    \begin{subfigure}[b]{0.49\linewidth}
    \centering
    \includegraphics[width=\linewidth]{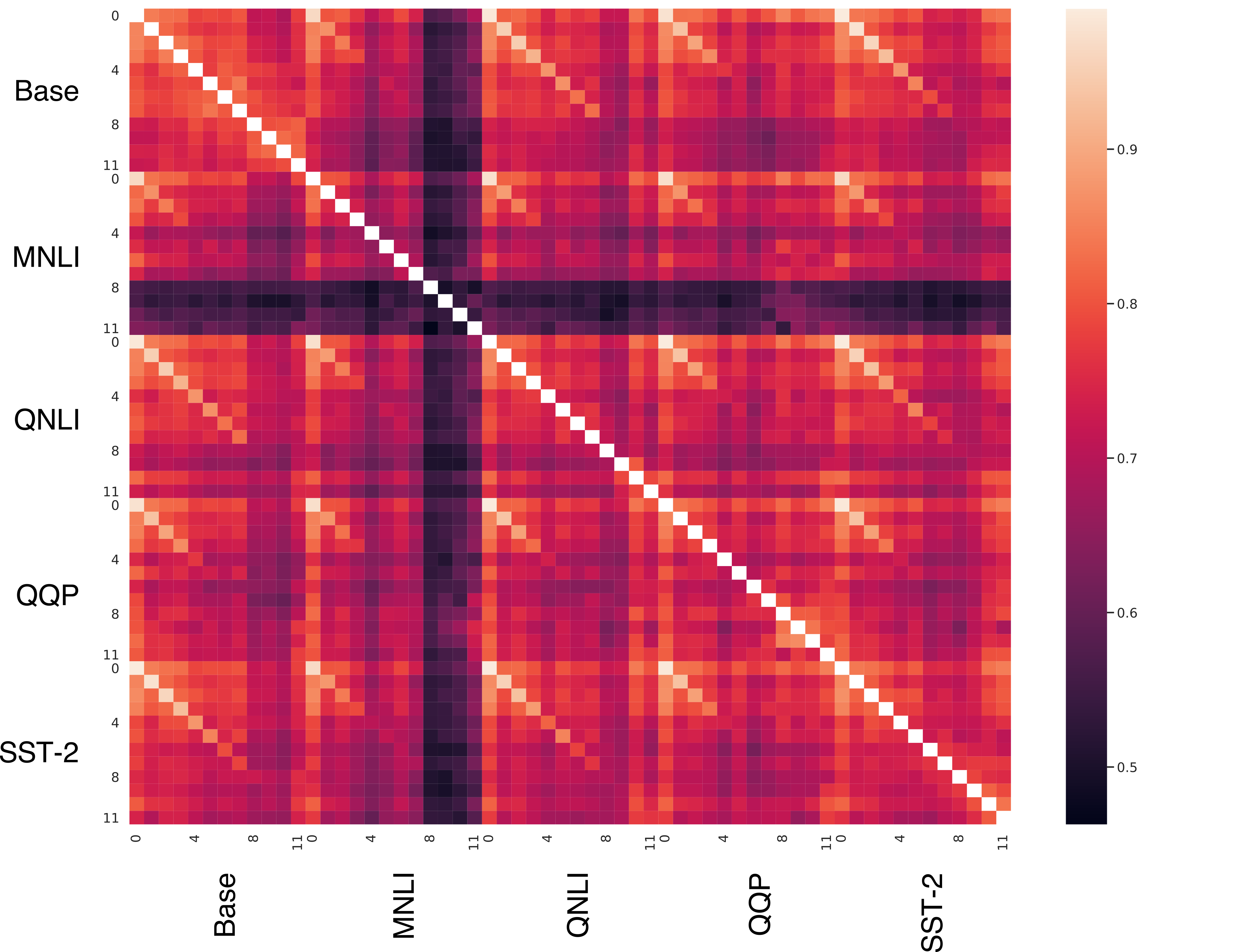}
    \caption{XLNet}
    \label{fig:fine-tuned-attn-heatmap-xlnet}
    \end{subfigure}
    \caption{Jensen--Shannon attention similarity heatmaps  of layers in base (pre-trained, not fine-tuned) and fine-tuned models.}
    \label{fig:fine-tuned-attn-heatmaps}
\end{figure*}

\subsection{Results}

\paragraph{Top layers are more affected by fine-tuning}
Figure~\ref{fig:fine-tuned-heatmaps} shows representation-level \linckasim~similarity heatmaps of pre-trained (not fine-tuned) and fine-tuned versions of BERT and XLNet. 
The most striking pattern is that the top layers are more affected by fine-tuning than the bottom layers, as evidenced by the low similarity of high layers of the pre-trained models with their fine-tuned counterparts. \citet{hao-etal-2019-visualizing} also observed that lower layers of BERT are less affected by fine-tuning than top layers, by visualizing the training loss surfaces.\footnote{A reviewer commented that this pattern seems like a natural consequence of back-propagation, which we concur with, although in on-going work we found that middle layers of XLNet lead to more gains when fine-tuned. Future work can also explore the effect of optimization on the similarity measures.} 
In Appendix~\ref{sec:efficient}, we demonstrate that this observation can motivate a more efficient fine-tuning process, where some of the layers are frozen while others are fine-tuned. 

There are some task-specific differences. 
In BERT, the top layers of the SST-2-fine-tuned model are affected more %
than other layers. This may be because SST-2 is a sentence classification task, while the other tasks are sentence-pair classification. A potential implication of this is that non-SST-2 tasks can contribute to one another in a multi-task fine-tuning setup. In contrast, in XLNet, fine-tuning on any task leads to top layers being very different from all layers of models fine-tuned on other tasks. This suggests that XLNet representations become very task-specific, and thus multi-task fine-tuning may be less effective with XLNet than with BERT.

\begin{figure*}[t]
    \centering
    \begin{subfigure}[b]{0.49\linewidth}
    \centering
    \includegraphics[width=\linewidth]{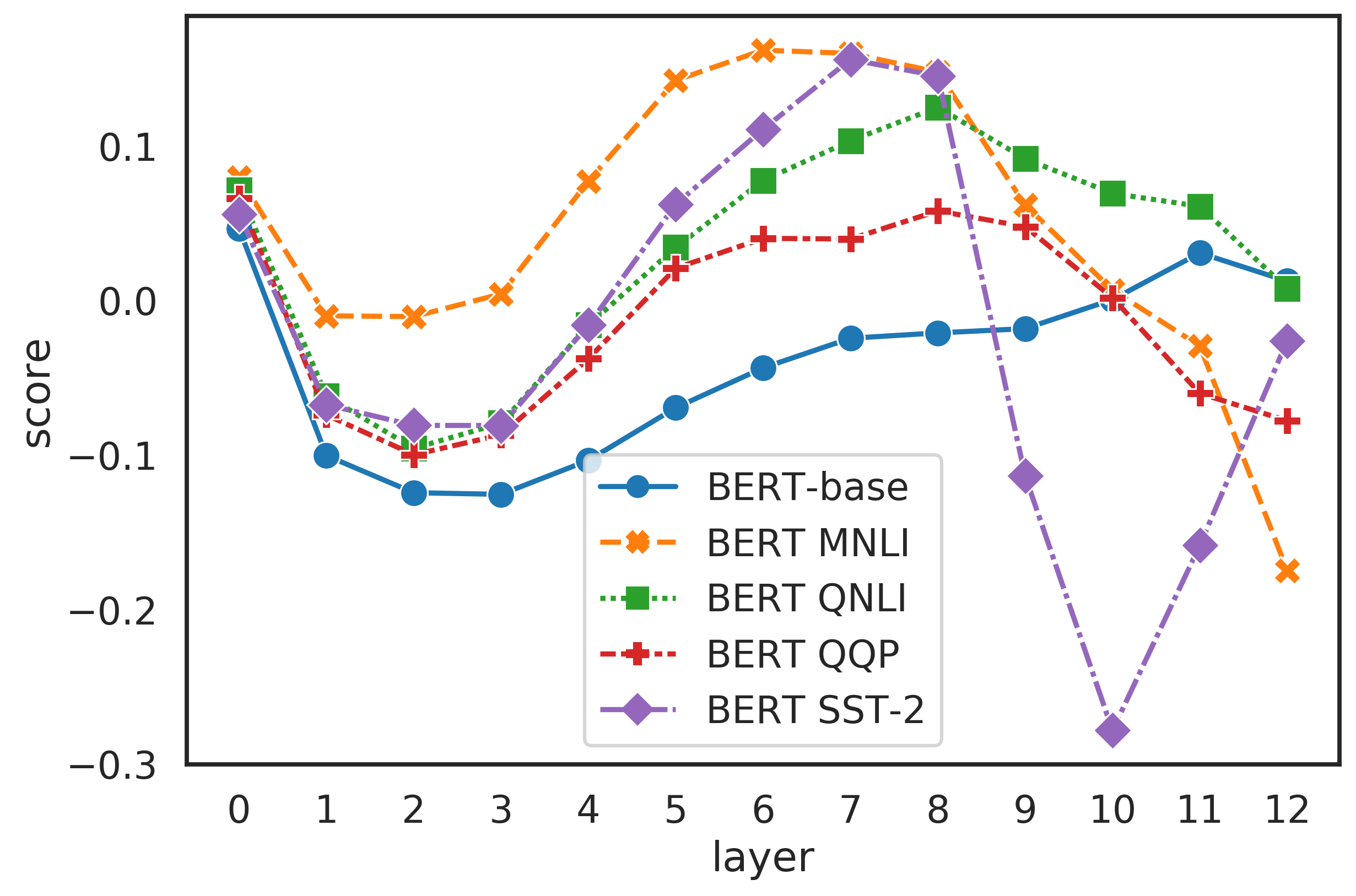}
    \caption{BERT}
    \label{fig:fine-tuned-loc-score-BERT}
    \end{subfigure} 
    \begin{subfigure}[b]{0.49\linewidth}
    \centering
    \includegraphics[width=\linewidth]{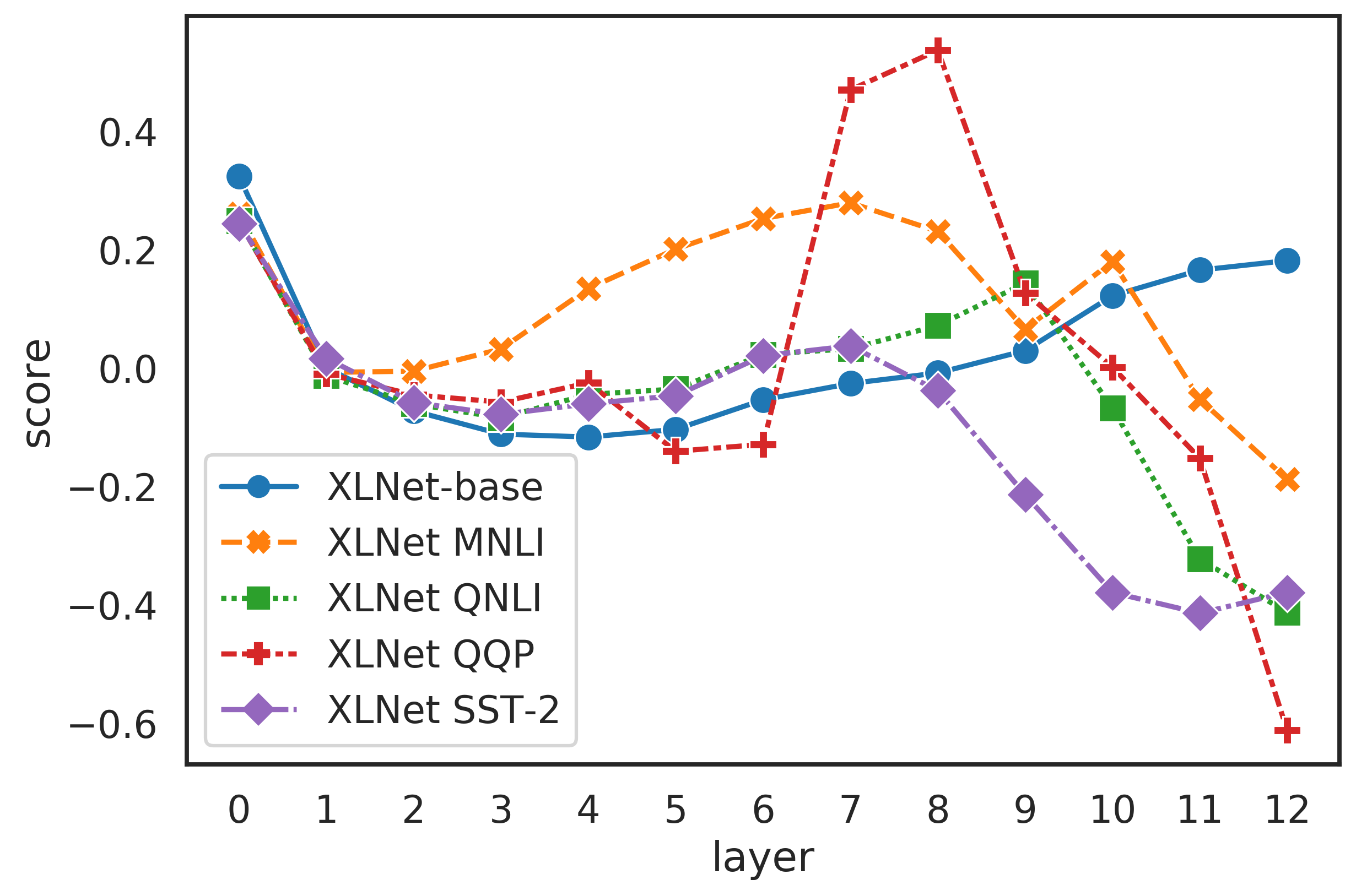}
    \caption{XLNet}
    \label{fig:fine-tuned-loc-score-xlnet}
    \end{subfigure}
    \caption{Localization scores per layer in base and fine-tuned models.}
    \label{fig:fine-tuned-loc-score}
\end{figure*}

Observing the \attnsim~similarity based on Jensen--Shannon divergence for base and fine-tuned models (Figure~\ref{fig:fine-tuned-attn-heatmaps}), we again see that top layers have lower similarities, implying that they undergo greater changed during fine-tuning. Other attention-based measures behaved similarly (not shown). 
\citet{kovaleva-etal-2019-revealing} made a similar observation by comparing the cosine similarity of attention matrices in BERT, although they did not perform cross-task comparisons. In fact, the diagonals within each block indicate that bottom layers remain similar to one another even when fine-tuning on different tasks, while top layers diverge after fine-tuning. 
The vertical bands at layers 0 mean that many higher layers have a head that is very similar to a head from the first layer, that is, a form of redundancy, which can explain why many heads can be pruned~\cite{michel2019sixteen,voita-etal-2019-analyzing,kovaleva-etal-2019-revealing}.  
Comparing BERT and XLNet, the vertical bands at the top layers of BERT (especially in MNLI, QQI, and SST-2) suggest that some top layers are very similar to any other layer. In XLNet, top MNLI layers are quite different from any other layer. Thus different objective functions impact the attention heads differently under fine-tuning.

\paragraph{Fine-tuning affects localization} 
Figure~\ref{fig:fine-tuned-loc-score} shows localization scores for different layers in pre-trained and fine-tuned models. In contrast to the pre-trained models, the fine-tuned ones decrease in localization at the top layers. This decrease may be the result of top layers learning %
high-level tasks, which require multiple neurons to capture properly.

\section{Conclusion}

In this work, we analyzed various prominent contextual word representations from the perspective of similarity analysis. We compared different layers of pre-trained models using both localized and distributed measures of similarity, at neuron, representation, and attention levels. We found that different architectures often have similar internal representations, but differ at the level of individual neurons. We also observed that higher layers are more localized than lower ones.  
Comparing fine-tuned and pre-trained models, we found that higher layers are more affected by fine-tuning in their representations and attention weights, and become less localized.  
These findings motivated experimenting with layer-selective fine-tuning, where we were able to obtain good performance while freezing the lower layers and only fine-tuning the top ones. 

Our approach is complementary to the linguistic analysis of models via probing classifiers. An exciting direction for future work is to combine the two approaches in order to identify which linguistic properties are captured in model components that are %
similar to one another, or explicate how localization of information contributes to the learnability of particular properties. 
It may be insightful to compare the results of our analysis to the loss surfaces of the same models, especially before and after fine-tuning~\cite{hao-etal-2019-visualizing}. One could also study whether a high similarity entail that two models converged to a similar solution. 
Our localization score can also be compared to other aspects of neural representations, such as gradient distributions and their relation to memorization/generalization~\cite{pmlr-v70-arpit17a}.
Finally, the similarity analysis may also help %
improve model efficiency, for instance by pointing to components that do not change much during fine-tuning and can thus be pruned. %

\section*{Acknowledgements}
We thank Nelson Liu for providing some of the representations analyzed in this work.  We also thank the anonymous reviewers for their many valuable comments. 
This research was carried out in collaboration between the HBKU Qatar Computing Research Institute (QCRI) and the MIT Computer Science and Artificial Intelligence Laboratory (CSAIL). Y.B.\ is also supported by the Harvard Mind, Brain, and Behavior Initiative (MBB).

\bibliography{acl2020,anthology}
\bibliographystyle{acl_natbib}

\clearpage

\appendix

\section{Mathematical Details of Similarity Measures}
\label{app:math}
We assume a fixed corpus with $W = \sum_i W_i$ total words, and $W^{(2)} = \sum_i W_i^2$ total pairs. Here $W_i$ is the number of words in sentence $i$. 

A representational layer $\wordrepr_{l}^{(m)}$ may be seen as a $W\times N_m$ matrix, where $N_m$ is the number of neurons (per layer) in model $m$. A single neuron $\wordrepr_{l}^{(m)}[k]$ (really $\wordrepr_{l}^{(m)}[:,k]$) is a $W\times 1$ column vector. 

An attention head $\attnhead_{l}^{(m)}[k]$ may be seen as a random variable ranging over sentences $s_i$ and taking matrix values $\attnhead_{l}^{(m)}[k](s_i) \in \mathbb{R}^{t_i\times t_i}$, %
$t_i = \len(s_i)$. 

\subsection{Neuron-level similarity}
For a given neuron $\wordrepr_{l}^{(m)}[k]$, we define 
\begin{align*}
    &\neuronsim(\wordrepr^{(m)}_{l}[k], \wordrepr^{(m')}_{l'}) = \\ \nonumber
    &\max_{k'} | \rho( \wordrepr^{(m')}_{l'}[k'], \wordrepr^{(m)}_{l}[k] ) |
\end{align*}
as the maximum correlation between it and another neuron in some layer~\cite{bau:2019:ICLR}. Here $\rho$ is the Pearson correlation. 
This naturally gives rise to an aggregate measure at the layer level: 
\begin{align*}
    \neuronsim(\wordrepr^{(m)}_l, \wordrepr^{(m')}_{l'}) = \\
    \frac{1}{N_m}\sum_k \neuronsim(\wordrepr^{(m)}_{l}[k], \wordrepr^{(m')}_{l'})
\end{align*}

\subsection{Mixed neuron--representation similarity}
We define
\begin{align*}
    &\mixedsim(\wordrepr_{l}^{(m)}[k], \wordrepr^{(m')}_{l'}) :=\\ \nonumber
    &\rval{\lstsq(\wordrepr_{l'}^{(m')}, \wordrepr_{l}^{(m)}[k])}
\end{align*}
where $\rval{}$ is the r-value associated with the regression, the norm of the prediction divided by the norm of the regressand. As before, this is extended to the layer level:
\begin{align*}
    \mixedsim(\wordrepr^{(m)}_l, \wordrepr^{(m')}_{l'}) = \\
    \frac{1}{N_m}\sum_k \mixedsim(\wordrepr^{(m)}_{l}[k], \wordrepr^{(m')}_{l'})
\end{align*}

\subsection{Representation-level similarity}
In the following, let $\bmat{Z}$ denote a column centering transformation. For a given matrix $\bmat{A}$, the sum of each column in $\bmat{Z}\bmat{A}$ is zero. 
\paragraph{SVCCA }  Given two layers 
\begin{align*}
    \bmat{X}, \bmat{Y} = \bmat{Z}\wordrepr_{l_x}^{(m_x)}, \bmat{Z}\wordrepr_{l_y}^{(m_y)}
\end{align*}
we compute the truncated principal components
\begin{align*}
    \bmat{X'}, \bmat{Y'} = \bmat{U_x}[:,:l_x], \bmat{U_y}[:,:l_y]
\end{align*}
where $\bmat{U_x}$ are the left singular vectors of $\bmat{X}$, and $l_x$ is the index required to account for 99\% of the variance. $\bmat{U_y}$ and $l_y$ are defined analogously. The SVCCA correlations, $\rho_{SVCCA}$, are defined as:
\begin{align*}
    \bmat{u}, \rho_{SVCCA}, \bmat{v} = \SVD(\bmat{X'}^T\bmat{Y'})
\end{align*}
The SVCCA similarity, $\svsim(\wordrepr_{l_x}^{(m_x)}, \wordrepr_{l_y}^{(m_y)})$, is the mean of $\rho_{SVCCA}$.

\paragraph{PWCCA }  Identical to SVCCA, except the computation of similarity is a weighted mean. Using the same notation as above, we define canonical vectors, 
\begin{align*}
    \bmat{H_X} := \bmat{X'}\bmat{u}\\
    \bmat{H_Y} := \bmat{Y'}\bmat{v}
\end{align*}
We define alignments
\begin{align*}
    \bmat{A_X} := \abs\left(\bmat{H_X^T}\bmat{X}\right)\\
    \bmat{A_Y} := \abs\left(\bmat{H_Y^T}\bmat{Y}\right)
\end{align*}    
where $\abs$ is the element-wise absolute value. The weights are
\begin{align*}
    \alpha_x := \weights(\bmat{A_X}\ones), \quad
    \alpha_y := \weights(\bmat{A_Y}\ones)
\end{align*}
where $\ones$ is the column vector of all ones, and $\weights$ normalizes a vector to sum to 1. The PWCCA similarity is
\begin{align*}
    \pwsim(\wordrepr_{l_x}^{(m_x)}, \wordrepr_{l_y}^{(m_y)}) := \alpha_x^T \rho_{SVCCA}\\
    \pwsim(\wordrepr_{l_y}^{(m_y)}, \wordrepr_{l_x}^{(m_x)}) := \alpha_y^T \rho_{SVCCA}
\end{align*}
It is asymmetric. 

\paragraph{CKA } We use the same notation as above. Given two layers,
\begin{align*}
    \bmat{X}, \bmat{Y} = \bmat{Z}\wordrepr_{l_x}^{(m_x)}, \bmat{Z}\wordrepr_{l_y}^{(m_y)}
\end{align*}
the CKA similarity is
\begin{align*}
    \linckasim(\wordrepr_{l_x}^{(m_x)}, \wordrepr_{l_y}^{(m_y)})
    := \frac{\norm{\bmat{X}^T\bmat{Y}}^2}{\norm{\bmat{X}^T\bmat{X}}\norm{\bmat{Y}^T\bmat{Y}}}
\end{align*}
where $\norm{\cdot} $ is the Frobenius norm. It is symmetric. 

\subsection{Attention-level similarity}
We define 
\begin{align*}
    &\attnsim(\attnhead^{(m)}_{l}[k], \attnhead^{(m')}_{l'}) = \nonumber \\ 
    &\max_{k'} \left[\Sim ( \attnhead^{(m')}_{l'}[k'], \attnhead^{(m)}_{l}[k] ) \right]
\end{align*}
We consider three such values of $\Sim$. 
\begin{itemize}[leftmargin=*,itemsep=2pt,parsep=2pt,topsep=3pt]
    \item Matrix norm: for each sentence $s_i$, compute the Frobenius norm
    $\norm{\attnhead^{(m')}_{l'}[h'](s_i) - \attnhead^{(m)}_{l}[h](s_i)}$. 
    Then average over sentences in the corpus. 
    \item Pearson correlation: for every word $x_i$, compare the attention distributions %
    the two heads induce from $x_i$ %
    to all %
    words under Pearson correlation: $\rho\left(\attnhead^{(m')}_{l',i}[h'], \attnhead^{(m)}_{l,i}[h]\right)$. Then average over words in the corpus.
    \item Jensen--Shannon divergence: for every word $x_i$, compare the attention distributions under Jensen--Shannon divergence:
    $\frac{1}{2}\KL(\alpha^{(m')}_{l',i}[h'] \bigm\| \beta) + \frac{1}{2}\KL(\alpha^{(m)}_{l,i}[h] \bigm\| \beta) $, where $\KL$ is the KL-divergence and $\beta$ is the average of the two attention distributions. Then average of words in the corpus. 
\end{itemize}
As before, this gives rise to aggregate measures at the layer level by averaging over heads $h$.

\section{Additional Representation-level Similarity Heatmaps}
\label{app:representation}

\begin{figure*}[t]
    \centering
    \begin{subfigure}[b]{0.49\linewidth}
    \centering
    \includegraphics[width=\linewidth]{figs/biglabels/lincka-biglabels-lowres.png}
    \caption{\linckasim}
    \label{fig:heatmap-cka}
    \end{subfigure}
    \begin{subfigure}[b]{0.49\linewidth}
    \centering
    \includegraphics[width=\linewidth]{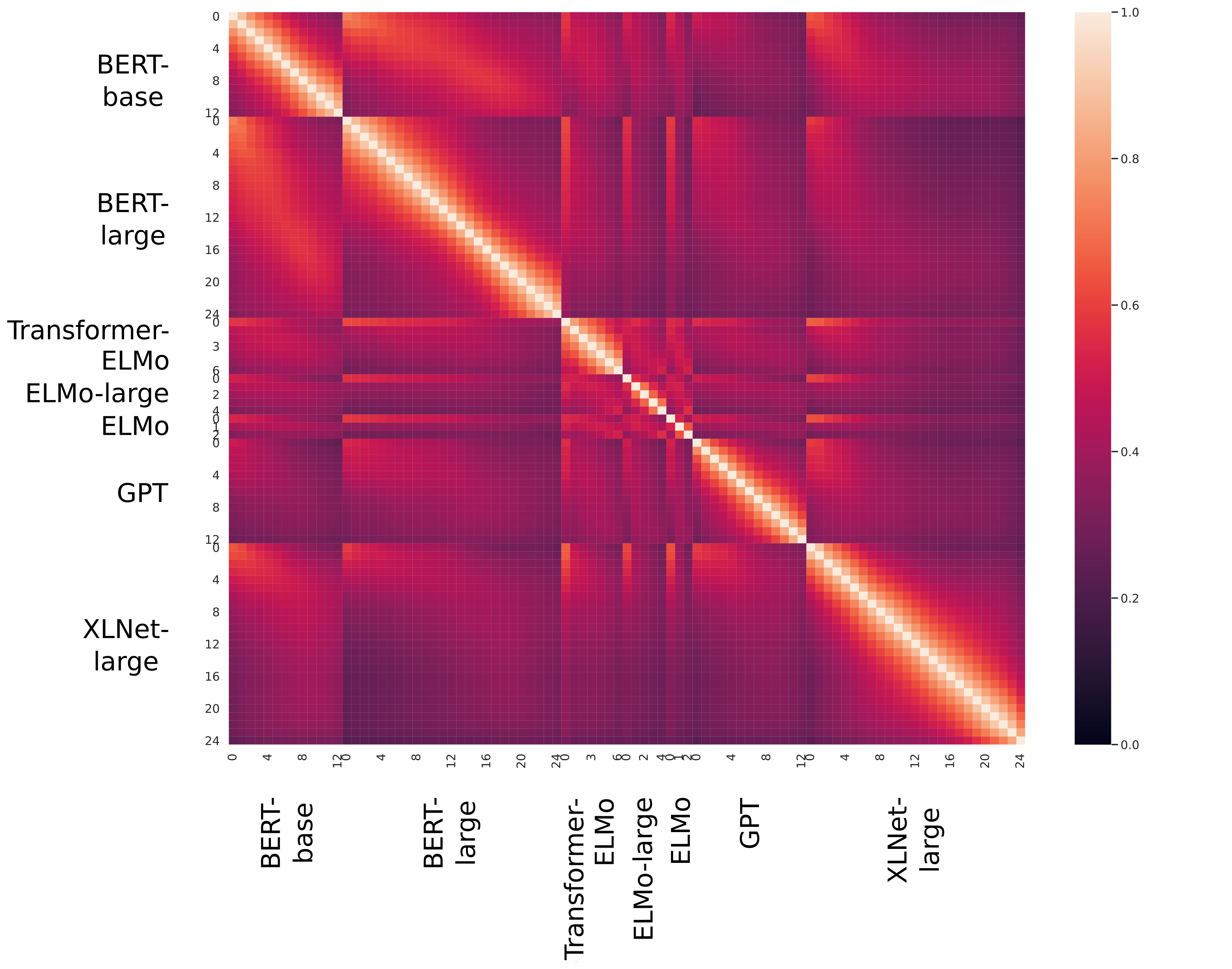}
    \caption{\svsim}
    \label{fig:heatmap-sv}
    \end{subfigure}\\
    \begin{subfigure}[b]{0.49\linewidth}
    \centering
    \includegraphics[width=\linewidth]{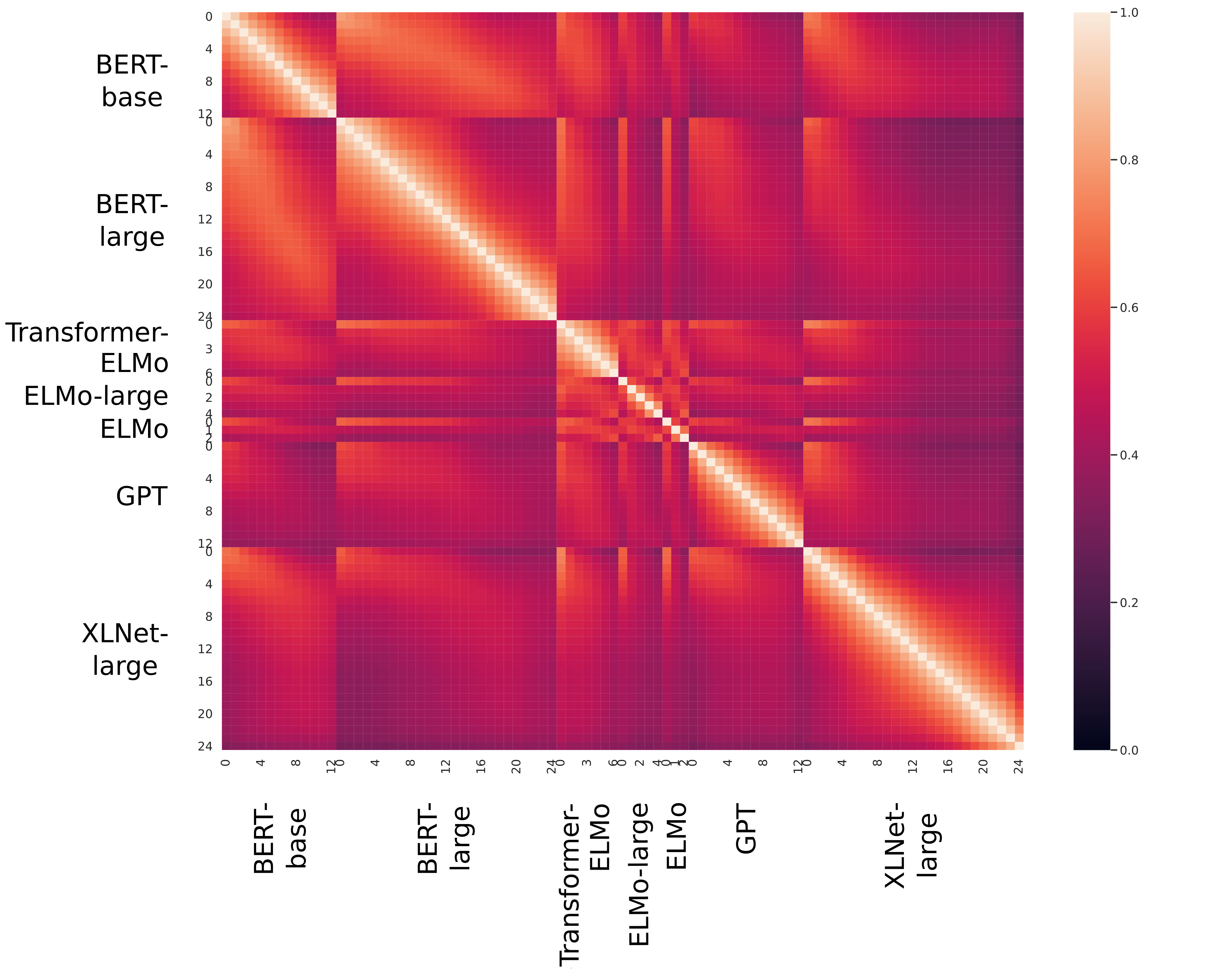}
    \caption{\pwsim}
    \label{fig:heatmap-pw}
    \end{subfigure}   
    \begin{subfigure}[b]{0.49\linewidth}
    \centering
    \includegraphics[width=\linewidth]{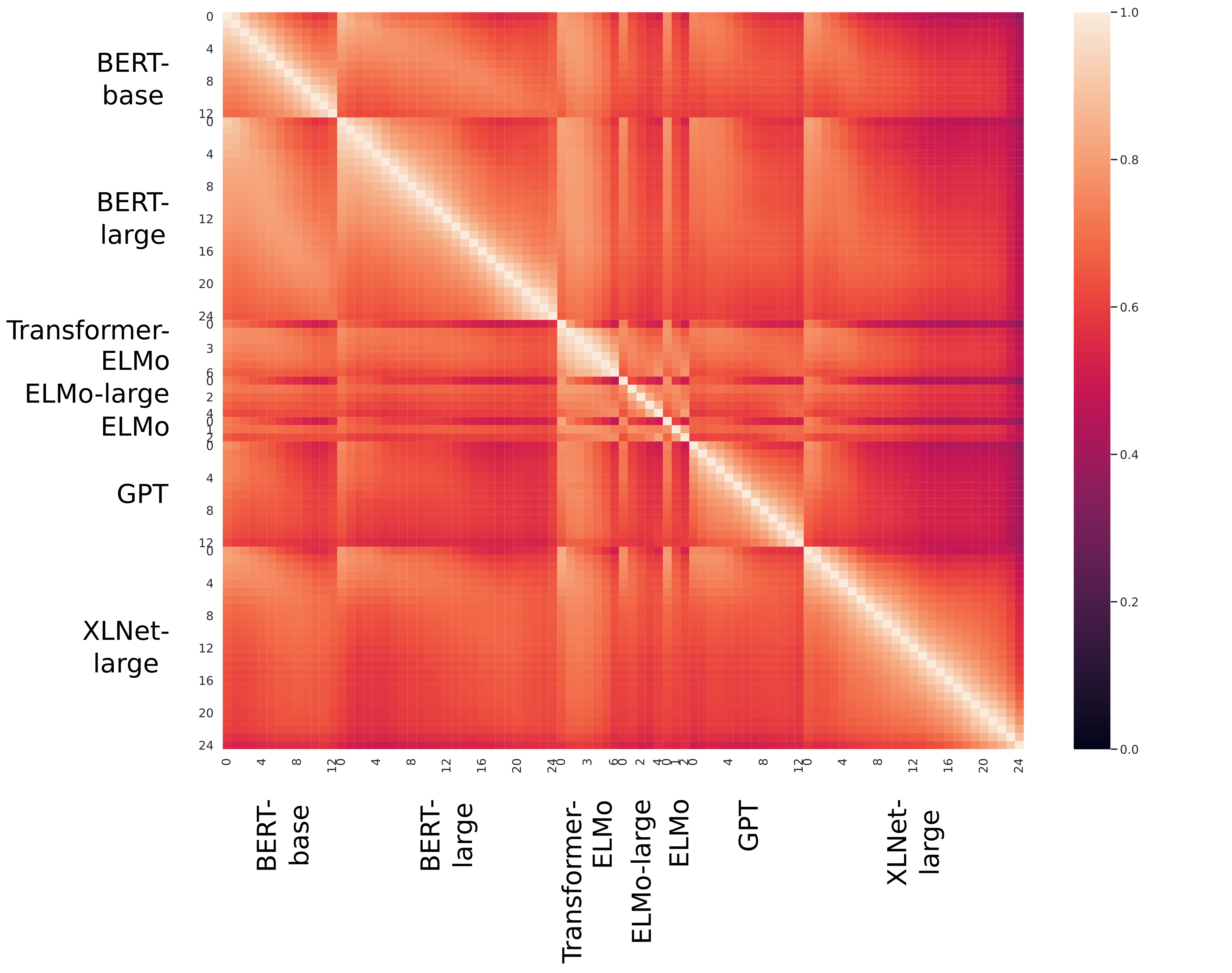}
    \caption{\mixedsim}
    \label{fig:heatmap-linreg}
    \end{subfigure}    
    \caption{Similarity heatmaps  of layers in various models under different %
    representation-level similarity measures.}
    \label{fig:heatmaps-all}
\end{figure*}

Figure~\ref{fig:heatmaps-all} shows additional representation-level similarity heatmaps. 

\subsection{Effect of Data Used for Similarity Measures} \label{app:altdata}

\begin{figure*}[t]
    \centering
    \begin{subfigure}[b]{0.49\linewidth}
    \centering
    \includegraphics[width=\linewidth]{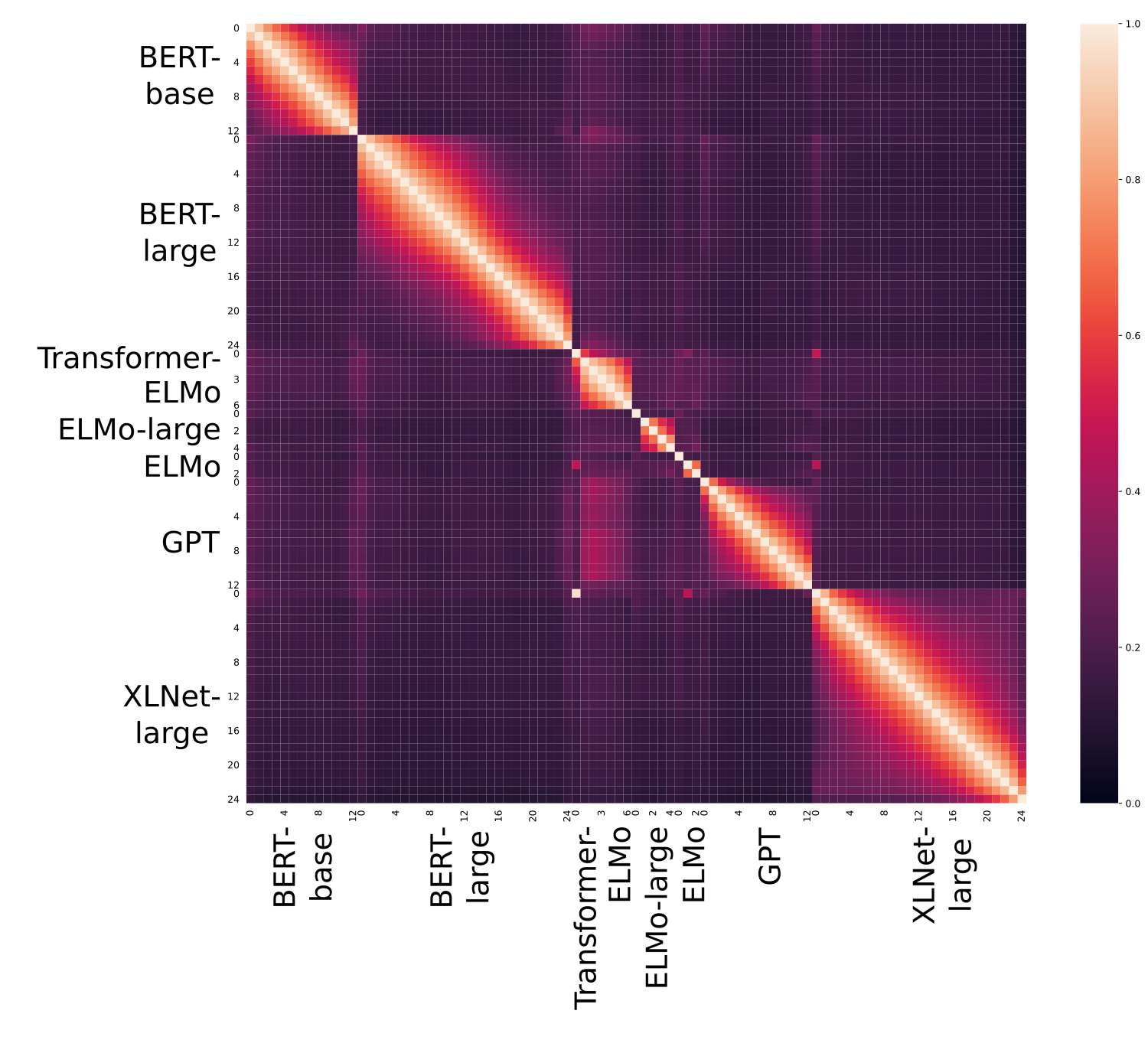}
    \caption{\neuronsim}
    \label{fig:heatmap-maxcorr-altdata}
    \end{subfigure}\\ 
    \begin{subfigure}[b]{0.49\linewidth}
    \centering
    \includegraphics[width=\linewidth]{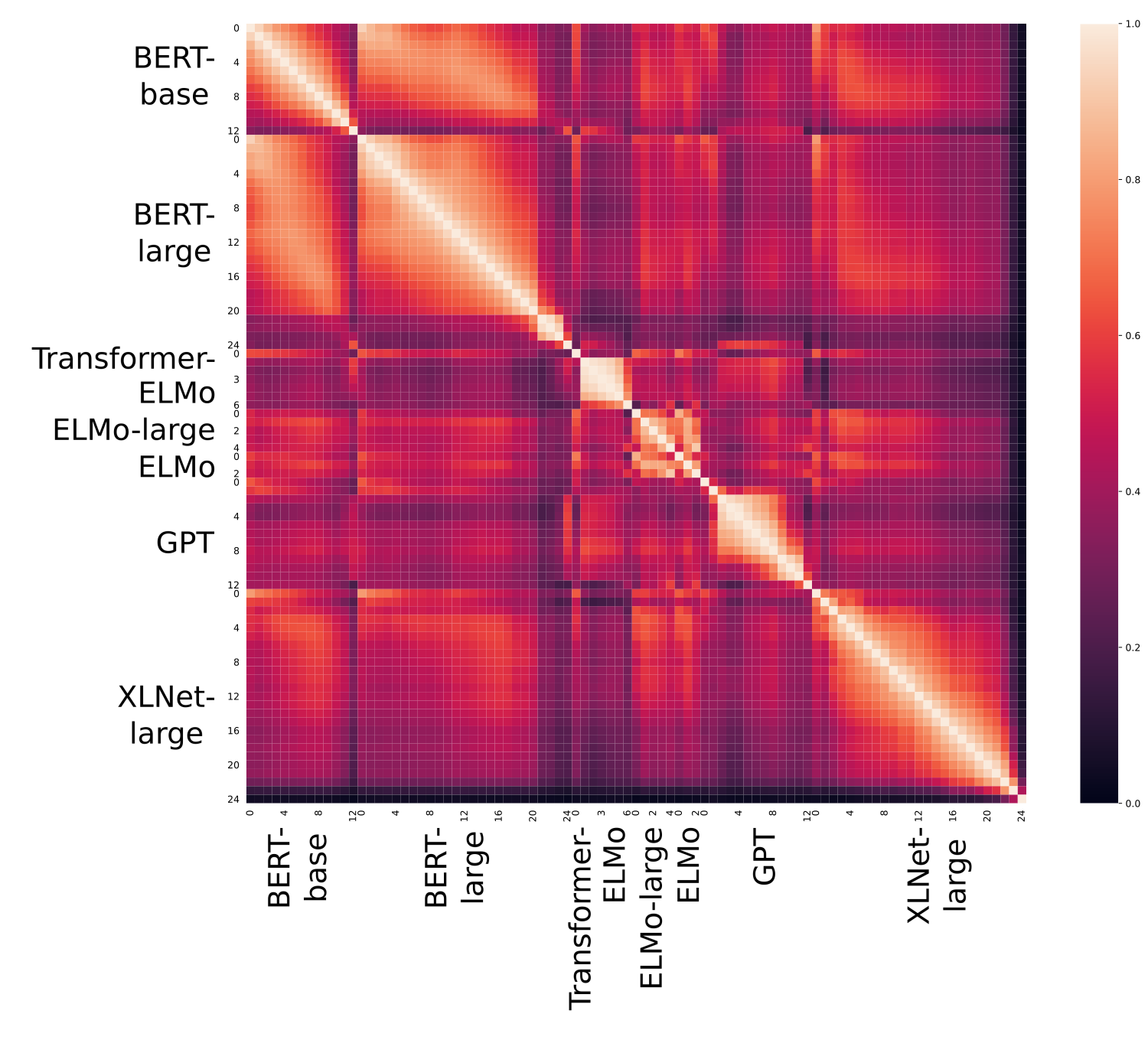}
    \caption{\linckasim}
    \label{fig:heatmap-cka-altdata}
    \end{subfigure}
    \begin{subfigure}[b]{0.49\linewidth}
    \centering
    \includegraphics[width=\linewidth]{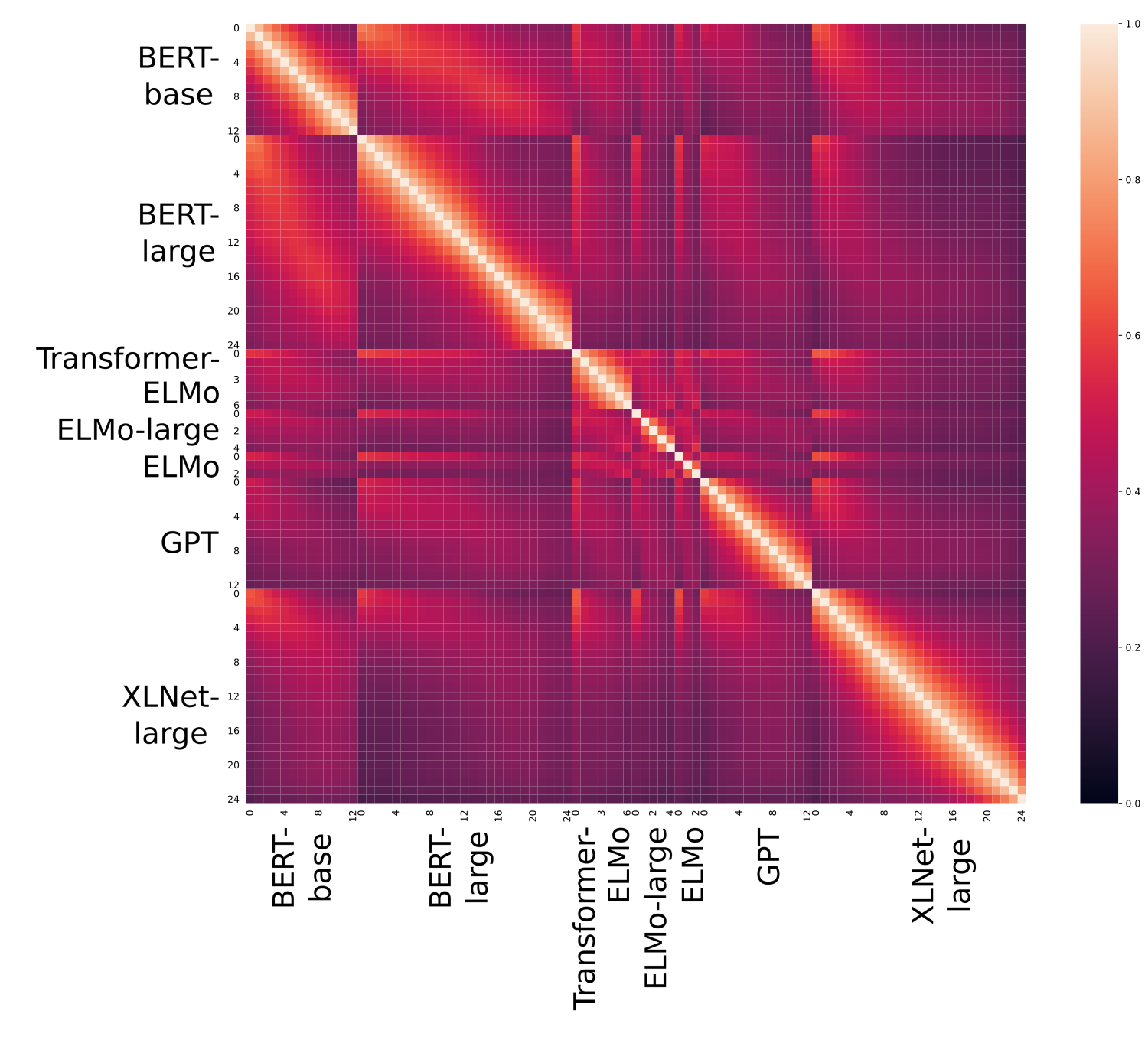}
    \caption{\svsim}
    \label{fig:heatmap-sv-altdata}
    \end{subfigure}\\
    \begin{subfigure}[b]{0.49\linewidth}
    \centering
    \includegraphics[width=\linewidth]{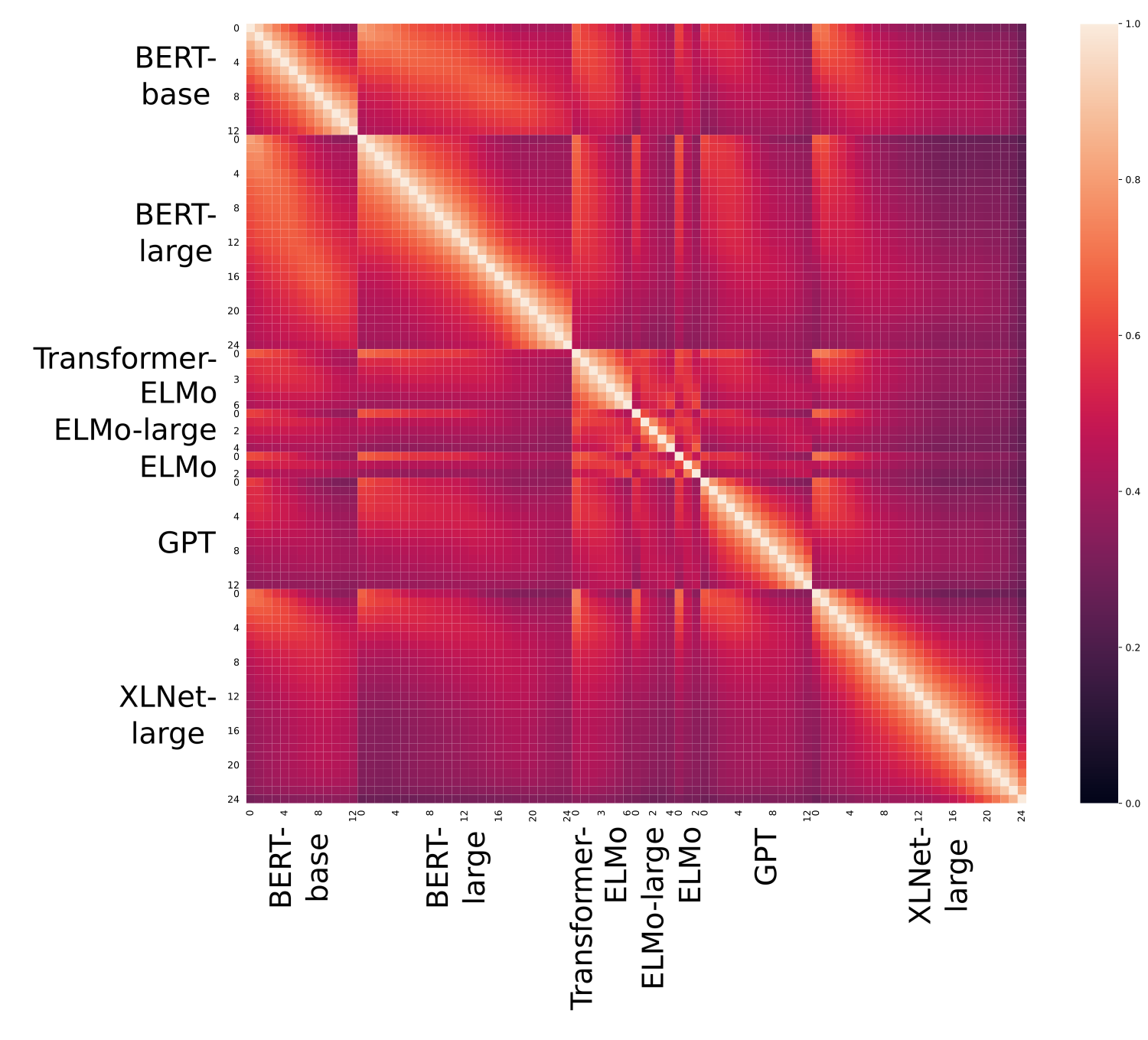}
    \caption{\pwsim}
    \label{fig:heatmap-pw-altdata}
    \end{subfigure}   
    \begin{subfigure}[b]{0.49\linewidth}
    \centering
    \includegraphics[width=\linewidth]{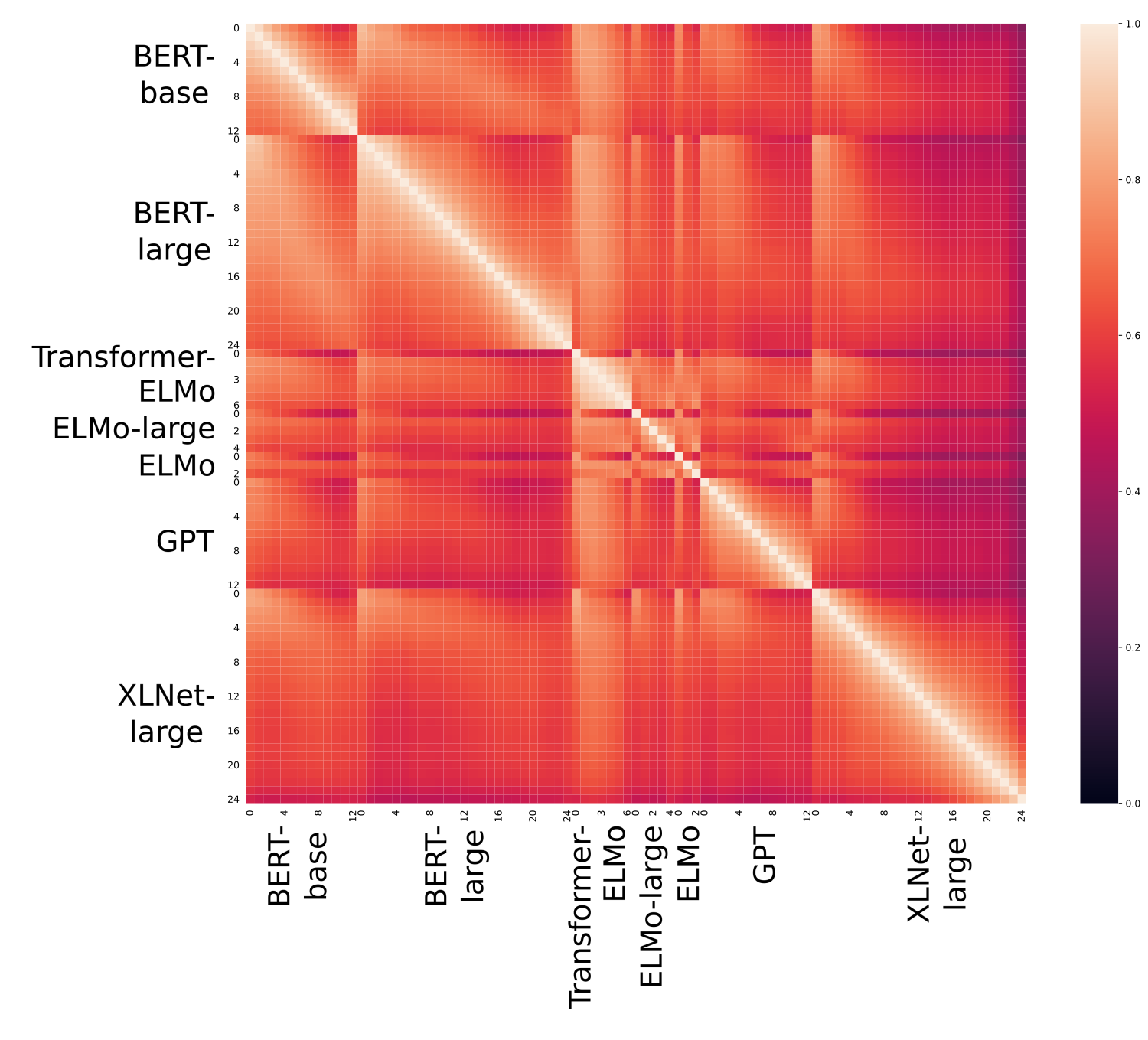}
    \caption{\mixedsim}
    \label{fig:heatmap-linreg-altdata}
    \end{subfigure}    
    \caption{Similarity heatmaps  of layers in various models under neuron-level and 
    representation-level similarity measures, using the English Web Treebank corpus.}
    \label{fig:heatmaps-all-altdata}
\end{figure*}

The majority of the experiments reported in the paper are using the Penn Treebank for calculating the similarity measures. Here we show that the results are consistent when using a different dataset, namely the Universal Dependencies English Web Treebank~\cite{silveira-etal-2014-gold}. We repeat the experiment reported in Section~\ref{sec:results-repr}. The resulting heatmaps, shown in Figure~\ref{fig:heatmaps-all-altdata}, are highly similar to those generated using the Penn Treebank, shown in Figure~\ref{fig:heatmaps-all}.

\section{Additional Attention-level Similarity Heatmaps} \label{app:attention}
\begin{figure*}[t]
    \centering
    \begin{subfigure}[b]{0.49\linewidth}
    \centering
    \includegraphics[width=\linewidth]{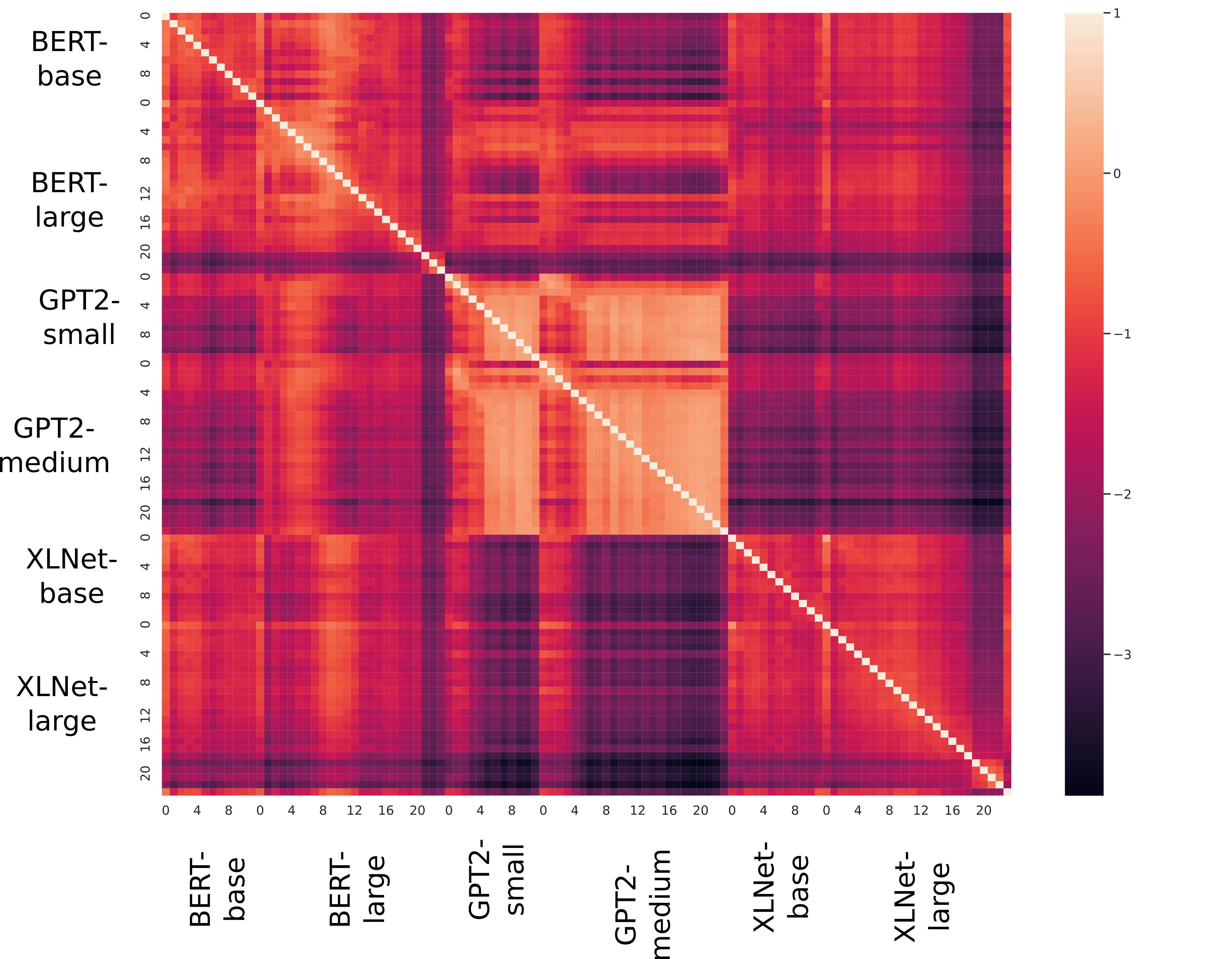}
    \caption{Matrix norm}
    \label{fig:heatmap-attn-matrixnorm}
    \end{subfigure}
    \begin{subfigure}[b]{0.49\linewidth}
    \centering
    \includegraphics[width=\linewidth]{figs/biglabels/attn-js-biglabels-lowres.png}
    \caption{Jensen--Shannon}
    \label{fig:heatmap-attn-js}
    \end{subfigure}\\    
    \begin{subfigure}[b]{0.49\linewidth}
    \centering
    \includegraphics[width=\linewidth]{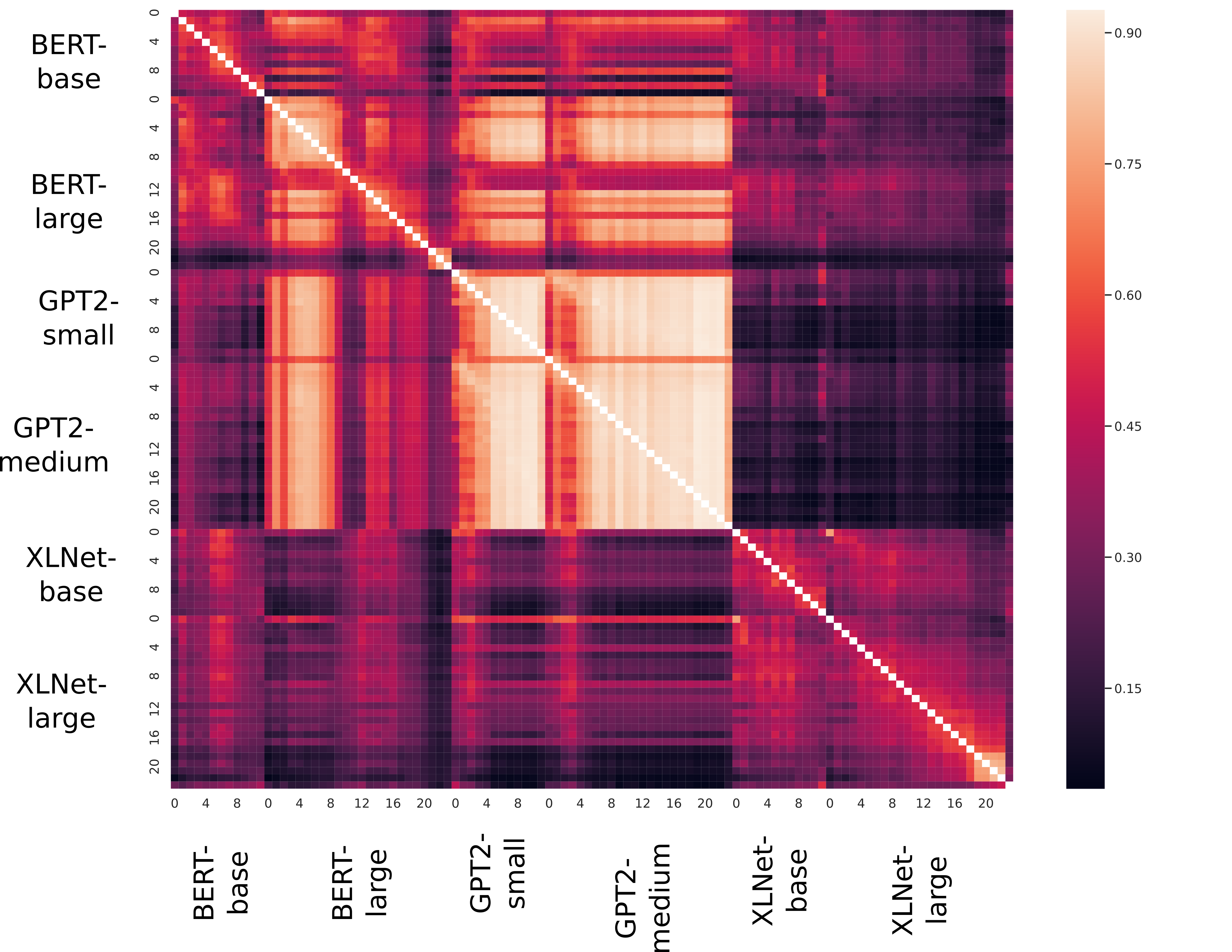}
    \caption{Pearson}
    \label{fig:heatmap-attn-pearson}
    \end{subfigure}
    \begin{subfigure}[b]{0.49\linewidth}
    \centering
    \includegraphics[width=\linewidth]{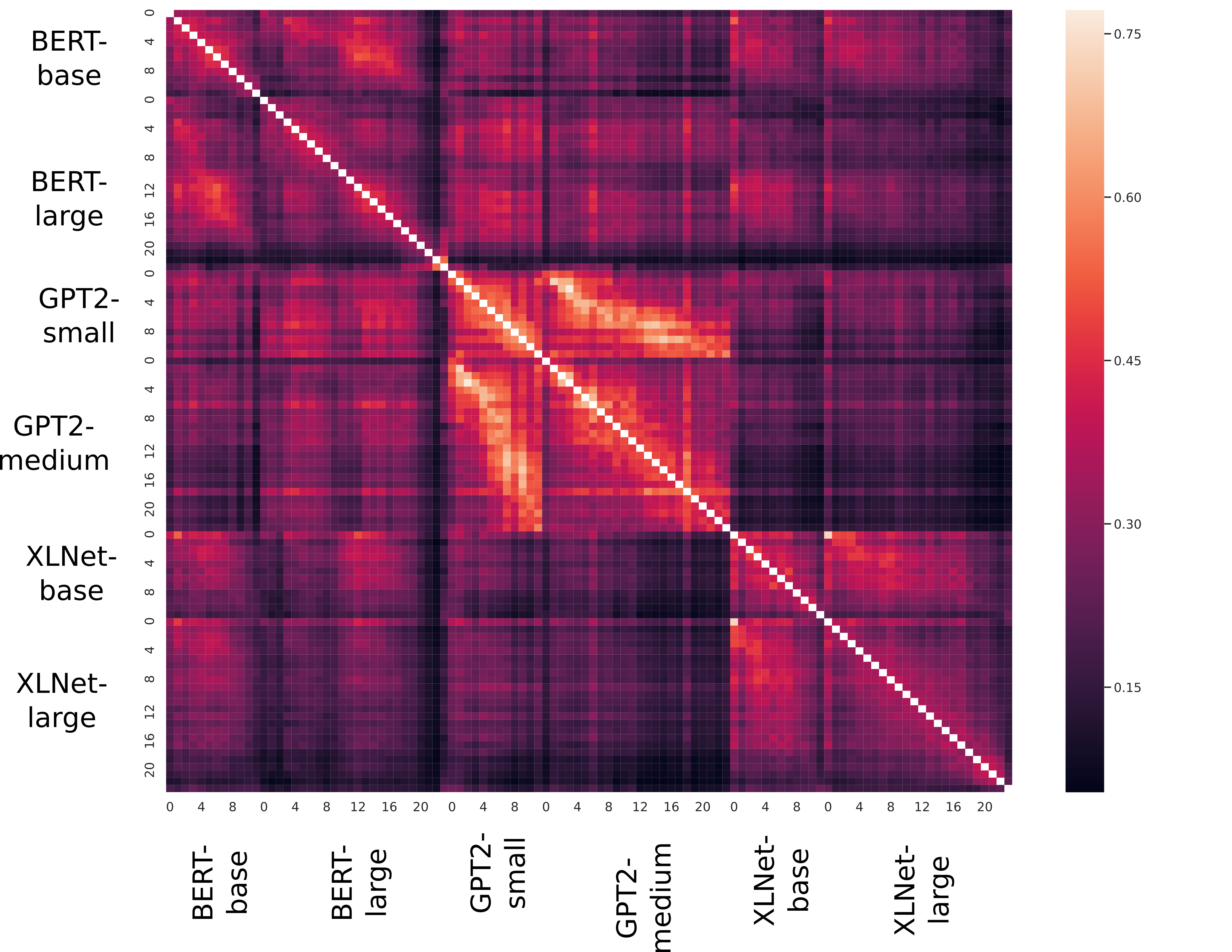}
    \caption{\svsim}
    \label{fig:heatmap-attn-svcca}
    \end{subfigure}\\    
    \begin{subfigure}[b]{0.49\linewidth}
    \centering
    \includegraphics[width=\linewidth]{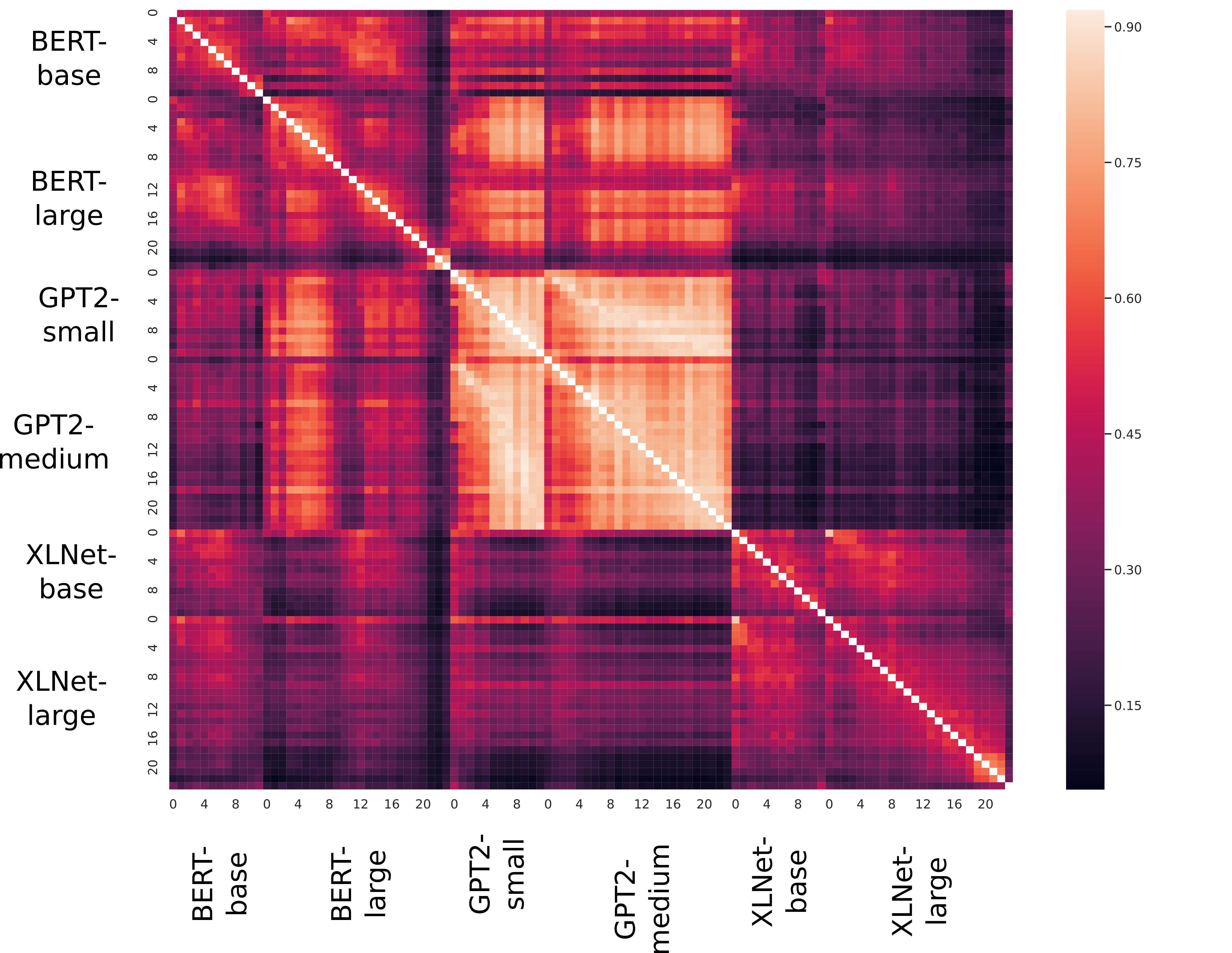}
    \caption{\pwsim}
    \label{fig:heatmap-attn-pwcca}
    \end{subfigure}    
    \begin{subfigure}[b]{0.49\linewidth}
    \centering
    \includegraphics[width=\linewidth]{figs/biglabels/attn-cka-biglabels-lowres.png}
    \caption{\linckasim}
    \label{fig:heatmap-attn-cka}
    \end{subfigure}  
    \caption{Similarity heatmaps  of layers in various models under different attention-level similarity measures.}
    \label{fig:attn-heatmaps-all}
\end{figure*}

Figure~\ref{fig:attn-heatmaps-all} shows additional attention-level similarity heatmaps. 

\section{Efficient Fine-tuning} \label{sec:efficient}
The analysis results showed that lower layers of the models go through limited changes during fine-tuning compared to higher layers. We use this insight to improve the efficiency of the fine-tuning process. In standard fine-tuning, back-propagation is done on the full network. We hypothesize that we can reduce the number of these operations by freezing the lower layers of the model since they are the least affected during the fine-tuning process. 
We experiment with freezing top and bottom layers of the network during the fine-tuning process. 
Different from prior work \cite{NIPS2017_7188,felbo-etal-2017-using,howard-ruder-2018-universal}, we freeze the selected layers for the complete fine-tuning process in contrast to freezing various layers for a fraction of the training time.
We use the default parameters settings provided in the Transformer library~\cite{Wolf2019HuggingFacesTS}: batch size = $8$, learning rate = $5e^{-5}$, Adam optimizer with epsilon = $1e^{-8}$, and number of epochs = 3. 

Table \ref{tab:bert_freezing_results} presents the results on BERT and XLNet. On all of the tasks except QQP, freezing the bottom layers resulted in better performance than freezing the top layers. One interesting observation is that as we increase the number of bottom layers for freezing to six, the performance marginally degrades while saving a lot more computation. Surprisingly, on SST-2 and QNLI, freezing the bottom six layers resulted in better or equal performance than not freezing any layers of both models. With freezing the bottom six layers, one can save back-propagation computation by more than 50\%. %

\begin{table}[t]
\centering
\begin{tabular}{l l llll}
\toprule
& Froze & SST-2 & MNLI & QNLI & QQP \\
\midrule
\multirow{5}{*}{\rotatebox[origin=c]{90}{BERT}} & 0 & 92.43 & 84.05 & 91.40 & 91.00 \\
 \midrule
 & Top 4 & 91.86 & 82.86 & 91.09 & \textbf{90.97} \\
& Bot. 4 & \textbf{92.43} & \textbf{84.16} & \textbf{91.85} & 90.86 \\
& Top 6 & 91.97 & 82.53 & 90.13 & 90.61 \\
& Bot. 6 & \textbf{93.00} & \textbf{84.00} & \textbf{91.80} & \textbf{90.71}  \\
\midrule
\multirow{5}{*}{\rotatebox[origin=c]{90}{XLNet}} & 0 & 93.92 & 85.97 & 90.35 & 90.55 \\
\midrule
 & Top 4 & 92.89 & 85.55 & 87.96 & \textbf{90.92} \\
& Bot. 4 & \textbf{93.12} & \textbf{86.04} & \textbf{90.65} & 89.36 \\
& Top 6 & 93.12 & 84.84 & 87.88 & \textbf{90.75} \\
& Bot. 6 & \textbf{93.92} & \textbf{85.64} & \textbf{90.99} & 89.02 \\
\bottomrule
\end{tabular}
\caption{Freezing top/bottom 4/6 layers of BERT and XLNet during fine-tuning.}
\label{tab:bert_freezing_results}
\vspace{-10pt}
\end{table}

\end{document}